\documentclass[letterpaper, 10 pt, conference]{ieeeconf}  % Comment this line out
\PassOptionsToPackage{hyphens}{url}

\IEEEoverridecommandlockouts                              % This command is only
\title{\LARGE \bf
Learning the Model While Learning Q: Finite-Time Sample Complexity of Online SyncMBQ
}

\usepackage{amsfonts}
\usepackage{amsmath,amsfonts,bm}

\def\eqref#1{equation~\ref{#1}}
\def\1{\bm{1}}

\def\eps{{\epsilon}}

\def\vb{{\bm{b}}}

\def\ve{{\bm{e}}}

\def\vp{{\bm{p}}}

\def\vv{{\bm{v}}}
\def\vw{{\bm{w}}}
\def\vx{{\bm{x}}}
\def\vy{{\bm{y}}}

\def\mA{{\bm{A}}}

\def\mD{{\bm{D}}}

\def\mI{{\bm{I}}}

\def\mP{{\bm{P}}}
\def\mQ{{\bm{Q}}}
\def\mR{{\bm{R}}}

\def\mPi{{\bm{\Pi}}}

\def\vpi{{\bm{\pi}}}

\DeclareMathAlphabet{\mathsfit}{\encodingdefault}{\sfdefault}{m}{sl}
\SetMathAlphabet{\mathsfit}{bold}{\encodingdefault}{\sfdefault}{bx}{n}

\def\gA{{\mathcal{A}}}

\def\gE{{\mathcal{E}}}

\def\gM{{\mathcal{M}}}

\def\gO{{\mathcal{O}}}
\def\gP{{\mathcal{P}}}

\def\gR{{\mathcal{R}}}
\def\gS{{\mathcal{S}}}

\def\sN{{\mathbb{N}}}

\def\sP{{\mathbb{P}}}

\newcommand{\E}{\mathbb{E}}

\newcommand{\R}{\mathbb{R}}

\usepackage{import}
\usepackage{algorithm} 
\usepackage{algpseudocode} 

\usepackage{graphicx}
\usepackage{subcaption}
\usepackage{xcolor}
\usepackage{xurl}
\usepackage{lipsum}
\usepackage{comment}

\newtheorem{theorem}{Theorem}
\newtheorem{remark}{Remark}
\newtheorem{lemma}{Lemma}

\usepackage{hyperref}
\usepackage{booktabs} 

\providecommand{\citep}[1]{\cite{#1}}
\providecommand{\citet}[1]{\cite{#1}}

\author{Han-Dogn Lim,  HyeAnn Lee and Donghwan Lee% <-this % stops a space
\thanks{H. Lim and D. Lee are with Korea Advanced Institute of Science and Technology (KAIST), Daejeon, Republic of Korea. A. Lee is with Samsung Electronics.  {\tt\small \{limaries30, donghwan\}@kaist.ac.kr, hyeann.lee@samsung.com}}%
\thanks{The full version of this paper is available at \url{https://arxiv.org/pdf/2402.11877}.}
}

\begin{document}

\maketitle
\thispagestyle{empty}
\pagestyle{empty}

%%%%%%%%%%%%%%%%%%%%%%%%%%%%%%%%%%%%%%%%%%%%%%%%%%%%%%%%%%%%%%%%%%%%%%%%%%%%%%%%
\begin{abstract}

    Reinforcement learning has witnessed significant advancements, particularly with the emergence of model-based approaches. Among these, $Q$-learning has proven to be a powerful algorithm in model-free settings. However, the extension of $Q$-learning to a model-based framework remains relatively unexplored. In this paper, we investigate the sample complexity of $Q$-learning when integrated with a model-based approach. The proposed algorihtms learns both the model and Q-value in an online manner. We demonstrate a near-optimal sample complexity result within a broad range of step sizes.

\end{abstract}

%%%%%%%%%%%%%%%%%%%%%%%%%%%%%%%%%%%%%%%%%%%%%%%%%%%%%%%%%%%%%%%%%%%%%%%%%%%%%%%%
\section{Introduction}

%Model-based reinforcement learning

Reinforcement learning (RL) aims to solve a sequential decision making problem. In a broad perspective, RL algorithms can be categorized into two classes: model-free and model-based approach. Both methods have shown great success in various scenarios~\citep{silver2017mastering,fawzi2022discovering}. The model-free approach tries to solve the sequential decision making process without any knowledge of a model. In contrast, the model-based approach leverages a known or estimated model during the learning process.

$Q$-learning, developed by~\cite{watkins1992q}, is one of the most widely used model-free RL algorithms. A rich body of literature has tried to understand the nature of $Q$-learning, and its non-asymptotic behavior has been recently understood in detail~\citep{qu2020finite,chen2019performance,lee2022finite,li2023q}. In particular,~\cite{li2023q} proved a tight sample complexity bound of $Q$-learning, which matches the lower bound on the sample complexity of $Q$-learning. %The {\color{blue}minimax sample complexity [this terminology is not normal so it would be better to remove minimax and instead explain its meaning in sentences]} of $Q$-learning has been recently established by~\cite{li2023q}, explaining the notorious sample inefficiency of model-free RL algorithms, which has been experimentally verified.

Meanwhile, a natural method to improve the sample efficiency of RL algorithms is to incorporate the model into the learning phase. Both in theoretical and experimental sense, leveraging the knowledge of learned or known model has been shown to improve over the model-free methods. For instance,~\cite{nagabandi2018neural,chua2018deep,kidambi2020morel} experimentally verified that approximating a model improves the sample efficiency of model-free algorithms. The sample efficiency of model-based algorithms has been studied in~\cite{kearns2002near,azar2012sample} under various settings in theoretical manners.

% In particular, model-based value iteration algorithm, $Q$-value iteration has been studied in~\cite{azar2012sample,kalathil2021empirical}, and~\cite{azar2012sample} showed that it can achieve minimax optimal sample complexity\footnote{We refer to minimax bound as the matching lower and upper bound of sample complexity in probable approximate correctness sense. }.

An intuitive method to extend $Q$-learning to a model-based approach, is to replace the stochastic components in the $Q$-learning update with estimators calculated via previous samples. In this paper, we study to what extent, we can improve the $Q$-learning with a model-based approach. We consider a natural extension of $Q$-learning with estimated model, which we call synchronous model-based $Q$-learning (SyncMBQ). In this respect, our main contributions are summarized as follows:
\begin{enumerate}
      \item We consider a new model-based RL algorithm, called SyncMBQ, which is an extension of $Q$-learning to model-based approaches. SyncMBQ can be seen as an online model-based $Q$-learning. The model estimate and $Q$-estimate are learned simultaneously in an online manner. It is online because the model and $Q$-estimate are updated in real-time using a single transition of the system at each iteration. We prove its near-optimal sample complexity under a general step-size regime, $\alpha\in (0,1]$ in terms of calculating $\eps$-optimal value function.
            % \item Moreover, SyncMBQ uses a more relaxed sampling model compared to the idealistic generative models (sampling oracle) used in some of previous model-based RLs. In particular, SyncMBQ uses a single transition at each iteration, where only one state-action pair is sampled from a fixed distribution under an exploratory behavior policy, and then the corresponding next state is sampled. The sampling method
            %       in our work is more realistic and weaker in the sense that the current state-action pair
            %       is only chosen from the given state-action distribution. This scenario aligns with the online learning setting, including weather forecast problem or sequential game scenario, where we cannot query
            %       any state-action pair in general.

      \item To the best of authors' knowledge, SyncMBQ considered in this paper has not been thoroughly investigated so far in the literature. In this paper, we prove that SyncMBQ can achieve $\tilde{\gO}\left( \frac{|\gS||\gA|}{(1-\gamma)^4\eps^2}  \right)$ sample complexity\footnote{$\tilde{\gO}(\cdot)$ hides the logarithmic factors.}, which is compatible with the optimal sample complexity $\tilde{\gO}\left( \frac{|\gS||\gA|}{(1-\gamma)^3\eps^2}  \right)$ achievable by existing model-based RLs with generative models except for the order of the effective horizon $\frac{1}{1-\gamma}$. In contrast to the generative models, we consider consider a relaxed sampling model, namely an i.i.d. sampling model. Moreover, this result improves the sample complexities achievable by existing model-based RLs without relying on generative models. In terms of online learning setting, our sample complexity is tighter than or comparable to the known bound achievable by model-based approaches.

            %Note also that this sample complexity is identical to the optimal sample complexity (not improvable) reported in~\cite{li2023q} for a model-free $Q$-learning with a restrictive step-size regime. SyncMBQ can achieve an identical efficiency under more general step-size regimes.

      \item For the analysis, the recently developed switching system model framework in~\cite{lee2022finite} is adopted in this paper. However, we develop a new switching system model for SyncMBQ and use it for our finite-time analysis.

      \item Finally, the performance of SyncMBQ is demonstrated via simulations, which empirically verify that SyncMBQ outperforms $Q$-learning under various scenarios.
\end{enumerate}

\noindent\textbf{Related Works on model-free Q-learning:}
Recently, some advances have been made on the non-asymptotic behavior of model-free $Q$-learning algorithms~\citep{qu2020finite, chen2019performance,lee2022finite,li2023q}.
For example, \cite{qu2020finite, chen2019performance,lee2022finite} studied asynchronous $Q$-learning, where only the $Q$-function estimate corresponding to a single state-action pair is updated at each iteration. Moreover, \cite{li2023q} proved that synchronous $Q$-learning, which updates the $Q$-function estimate for every state-action pairs each iteration, requires $\tilde{\gO}\left(\frac{|\gS||\gA|}{(1-\gamma)^4\eps^2}\right)$ number of samples to compute the $\eps$-optimal value function, which matches the known lower bound of the sample complexity of $Q$-learning. On the other hand,~\cite{dong2019q} studied $Q$-learning with upper confidence bound algorithm that achieve sample complexity of $\tilde{\gO}\left(\frac{|\gS||\gA|}{(1-\gamma)^7\eps^2}\right)$ to compute $\epsilon$-optimal policy\footnote{Computing an $\eps$-optimal policy and $\eps$-optimal value function requires different strategies. In this work, we focus on computing an $\eps$-optimal value function. A formal definition of $\eps$-policy can be found in page 101 in~\cite{kakade2003sample}.}.~\cite{zhang2021model} improved the bound to $\tilde{\gO}\left(\frac{|\gS||\gA|}{(1-\gamma)^3\eps^2}\right)$ for small range of $\eps$.

\noindent\textbf{Related works on model-based Q-learning:}
Model-based RL algorithms have been studied in~\cite{azar2012sample,gheshlaghi2013minimax,li2020breaking,agarwal2020model} under the assumption that generative models are available, where a generative model means a sampling oracle where we can access any state-action pair in the environment on our choice and get the next state. Roughly speaking, the main idea of these works is to run a $Q$-value iteration with the transition matrix and reward function replaced with the corresponding estimated models learned through the generative models. These model-based algorithms which consist of the two phases: in the first phase, the model is learned through samples from generative models; in the second phase, dynamic programmings are applied using the learned model. The authors of~\cite{azar2012sample} established a lower bound of $\tilde{\gO}\left(  \frac{|\gS||\gA|}{(1-\gamma)^3\eps^2} \right)$ on the sample complexity inherent to all RL algorithms and demonstrated that the sample complexity of their algorithm aligns with this established lower bound when a sufficient number of samples are collected in the first phase. More recently, \cite{li2020breaking} derived a similar sample complexity $\tilde{\gO}\left(  \frac{|\gS||\gA|}{(1-\gamma)^3\eps^2} \right)$ for $\eps$-optimal policy, and \cite{agarwal2020model} theoretically studied the optimality of these model-based $Q$-learning algorithms based on generative models.

Batch $Q$-learning algorithms, which use a number of samples collected at each iteration to construct an approximate Bellman operator, can be also interpreted as a non-parametric model-based $Q$-learning. Relying on an empirical mean estimator of the Bellman operator makes similar effect as leveraging a model estimate while reducing the space complexity from $\gO(|\gS|^2|\gA|)$ to $\gO\left(|\gS||\gA|\right)$. For example,~\cite{kearns1998finite,kalathil2021empirical,wainwright2019variance} studied empirical $Q$-value iteration algorithms, which estimate empirical Bellman operator with newly collected samples from a generative model at every iteration.
In particular, \cite{kearns1998finite} investigated the so-called phased $Q$-learning that uses estimated transition matrix estimated with newly collected samples at each iteration using the generative model. Phased $Q$-learning has been proven to achieve $\tilde{\gO}\left( \frac{|\gS||\gA|}{(1-\gamma)^4\eps^2} \right)$ sample complexity. This algorithm includes both a non-parametric batch version of $Q$-learning and a parametric model-based $Q$-learning version. In~\cite{kalathil2021empirical}, the authors proved that their empirical $Q$-value iteration can have $\tilde{\gO}\left(\frac{|\gS||\gA|}{(1-\gamma)^4\eps^2}\right)$ sample complexity to achieve desired level of accuracy. Moreover,~\cite{sidford2018near} proposed a slightly different variance-reduced $Q$-value iteration algorithm and obtained $\tilde{\gO}\left(\frac{|\gS||\gA|}{(1-\gamma)^3\eps^2} \right)$ sample complexities to compute an $\eps$-optimal policy. Afterwards,~\cite{wainwright2019variance} developed a variance-reduced $Q$-learning (a batch $Q$-learning), and proved that it can also achieve the optimal sample complexity $\tilde{\gO}\left( \frac{|\gS||\gA|}{(1-\gamma)^3\eps^2} \right)$. Note that all the aforementioned approaches require an access to a generative model, which is one of the most idealistic scenarios. The so-called delayed $Q$-learning, introduced by~\cite{strehl2006pac}, is also a batch $Q$-learning algorithm, which can be seen as a non-parametric model-based $Q$-learning. It has been proved that the delayed $Q$-learning can achieve $\tilde{\gO}\left(\frac{|\gS||\gA|}{(1-\gamma)^7\eps^4}\right)$ sample complexity, while it does not require a generative model, and uses samples from a single trajectory.

Without relying on generative models, more practical model-based RL algorithms have been studied in~\cite{kearns2002near,brafman2002r,strehl2005theoretical,szita2010model} with corresponding sample complexities, where explorations and single trajectory-based samples are used.
%where the sample complexity is measured in a slightly different manner: the number of samples to achieve an $\eps$-optimal policy.\footnote{This is known as sample complexity of exploration, and refer to~\cite{kakade2003sample} for more rigorous definition.} 
In particular, \cite{kearns2002near} proposed the so-called $E^3$ algorithm which decides to explore or exploit relying on the estimated model, and \cite{brafman2002r} introduced the so-called R-max algorithm wherein an agent acts by an optimal policy derived from the estimated model, and requires $\tilde{\gO}\left( \frac{|\gS|^2|\gA|}{(1-\gamma)^6\eps^3} \right)$ number of samples. In a subsequent work,~\cite{szita2010model} proposed a modified version of R-max algorithm that achieves $\tilde{\gO}\left(\frac{|\gS||\gA|}{(1-\gamma)^6\eps^2}\right)$ sample complexity. Moreover,~\cite{strehl2005theoretical} used a model based interval estimation method~\citep{wiering1998efficient} to deal with the exploration-exploitation dilemma and proved that it can also achieve $\tilde{\gO}\left( \frac{|\gS|^2|\gA|}{(1-\gamma)^6\eps^3} \right)$ sample complexity. In~\cite{lattimore2014near}, the authors studied the so-called UCRL$\gamma$ algorithm that achieves $\tilde{\gO}\left( \frac{|\gS|^2|\gA|}{(1-\gamma)^3\eps^2} \right)$ sample complexity, which is optimal in terms of the effective horizon $\frac{1}{1-\gamma}$, while sub-optimal in terms of the state size $|{\cal S}|$.

% Lastly, we note that~\cite{gheshlaghi2013minimax,sidford2018near,li2020breaking,agarwal2020model} studied sample complexities to obtain $\eps$-optimal policy. 
% %Obtaining $\eps$-optimal policy requires several more steps than deriving the $\eps$-optimal value function. 
% In this work, we focus on the sample complexity to obtain the $\eps$-optimal value function rather than $\eps$-optimal policy.

\section{Preliminaries}
\subsection{Markov decision process}

A Markov decision process (MDP) consists of five tuples $(\gS,\gA,\gamma,\gP,\gR)$ where $\gS:=\{1,2,\dots,|\gS| \}$ is the collection of states, $\gA:=\{1,2,\dots,|\gA| \}$ is the collection of actions, $\gamma\in (0,1)$ is the discount factor, $\gP:\gS\times\gA\times\gS \to [0,1]$ is the transition kernel, and $\gR:\gS\times\gA\times\gS\to\R$ is the reward function. Upon taking action $a\in\gA$ at state $s\in\gS$, transition to $s^{\prime}\in\gS$ occurs with probability $\gP(s,a,s^{\prime})$ and reward $\gR(s,a,s^{\prime})$ is incurred. For simplicity of the proof, we assume that the reward function is bounded, i.e., $|\gR(s,a,s^{\prime}) | \leq 1$ for all $(s,a,s^{\prime})\in\gS\times\gA\times\gS$.

A deterministic policy $\pi:\gS\to\gA$ maps a state $s\in\gS$ to an action $a\in\gA$. We study finding an optimal deterministic policy, $\pi^*$, that maximizes the expected sum of discounted rewards under an infinite horizon MDP setting, i.e.,
\begin{align*}
    \pi^* :=  \arg\max_{\pi\in\Omega}\sum_{k=0}^{\infty} \mathbb{E}\left[ \gamma^k r_k  \middle | \pi \right],
\end{align*}
where $\Omega$ is the set of all admissible deterministic policies, $r_k:=\gR(s_k,a_k,s_{k+1})$, $\{(s_k,a_k)\in\gS\times\gA\}_{k\in\sN}$ is a sequence of state-action trajectory generated by a Markov decision process under policy $\pi$, and  $\E\left[\cdot \middle | \pi\right]$ denotes the expected value conditioned on policy $\pi$. The $Q$-function under policy $\pi$, $Q^{\pi}:\gS\times\gA \to \R$, denotes expected cumulative discounted reward starting at $(s,a) \in  \gS\times \gA$ and following policy $\pi$ afterwards:
\begin{align*}
    Q^{\pi}(s,a):= \sum^{\infty}_{k=0} \E\left[ \gamma^k r_k \middle | (s_0,a_0)=(s,a),\;\pi\right].
\end{align*}

The optimal $Q$-function, which is a $Q$-function induced by the optimal policy $\pi^*$, is denoted as $Q^*(s,a):=Q^{\pi^*}(s,a)$ for all $s,a\in\gS\times\gA$. The optimal policy $\pi^*$ can be recovered once $Q^*$ is known, i.e., $\pi^*(s)=arg\max_{a\in\gA} Q^{*}(s,a)$. It is well known that the optimal $Q$-function, $Q^*$, satisfies the following equation, the so-called optimal Bellman equation~\citep{bellman1966dynamic}, for all $s\in\gS,a\in\gA$:
\begin{align*}
    Q^*(s,a) =  \sum_{s^{\prime}\in\gS}\gP(s,a,s^{\prime})\left(\gR(s,a,s^{\prime})+\gamma\max_{u\in\gA} Q^*(s,u)\right).
\end{align*}
% where the operator $\gB$ is defined on the $Q$-function $\mQ:\gS\times\gA\to\R$, 
% :\R^{|\gS||\gA|} \times \gS\times \gA \to \R $ is called the Bellman operator, and $\mQ^*=\begin{bmatrix}
%     \mQ^*(1,1) & \mQ^*(1,2) & \dots & \mQ^*(|\gS|,|\gA|)
% \end{bmatrix}^{\top} \in \R^{|\gS||\gA|}$.

\subsection{Overview of Q-learning}
% RL algorithms can be broadly classified into two categories : model-free and model-based approaches. Model-free algorithms~\citep{watkins1992q,sutton1999policy} learns optimal policy without any knowledge of a model, specifically, transition probability or the reward function. In contrast, model-based approach~\citep{kearns1998finite,kearns2002near} leverages a known or estimated model during the learning process.

In this section, we briefly illustrate the $Q$-learning algorithm~\citep{watkins1992q}. The $Q$-learning algorithm is one of the well-known model-free algorithms. The update of $Q$-learning can be written as, for $k\in \sN$, and $\mQ_0\in\R^{|\gS||\gA|}$:
\begin{align}
      & \mQ_{k+1}(s_k,a_k)                                                                                               \nonumber \\
    = & (1-\alpha) \mQ_k (s_k,a_k)+\alpha \bigl(r_k+\gamma \max_{a\in\gA} \mQ_k(s_{k+1},a)\bigr), \label{eq:model-free-Q}
\end{align}
where $\mQ_k\in\R^{|\gS||\gA|}$, $\mQ_k(s,a)=(\ve_{s}\otimes \ve_{a})^{\top}\mQ_k$ for $(s,a)\in\gS\times \gA$, $\alpha\in (0,1]$ is the constant step-size, and $\ve_s$ and $\ve_a$ denote $s$-th and $a$-th canonical basis vectors in $\R^{|\gS|}$ and $\R^{|\gA|}$, respectively.

% The algorithm is viewed as the model-free approach since it does not assume knowledge of transition kernel or the reward function for the entire learning process.

To proceed, we further assume that at time step $k$, the state-action pair $(s_k,a_k)$ is sampled from a fixed probability $d:\gS\times\gA \to [0,1]$. We will assume that $d(s,a)>0$ for all $s,a\in\gS\times\gA$, and let $d_{\min}:=\min_{s,a\in\gS\times\gA} d(s,a)$. Furthermore, we introduce a set of matrix notations. Let
$\mP\in \R^{|\gS||\gA|\times |\gS|}$ be the transition matrix whose $(s-1)|\gA|+a$-th row vector equals $(\vp^{s,a})^{\top}$, where $\vp^{s,a} \in \R^{|\gS|}$ is a vector with entries $[\vp^{s,a}]_{s^{\prime}}=\gP(s,a,s^{\prime})$. Let $\mR\in\R^{|\gS||\gA|}$ denote the expected reward vector, whose $(s-1)|\gA|+a$-th element corresponds $\E\left[r(s,a,s^{\prime})\middle | s,a\right]$. Define $\mD\in\R^{|\gS||\gA|}$ to be a diagonal matrix such that its $(s-1)|\gA|+a$-th element corresponds to $d(s,a)$. Moreover, for any deterministic policy $\pi:\gS\to\gA$,  define, $\mPi^{\pi}\in \R^{|\gS|\times |\gS||\gA|}$ to be a matrix such that its $s$-th row vector corresponds to $\ve_s^{\top}\otimes \vpi(s)^{\top}$ for $s\in\gS$ where $\ve_s \in \R^{|\gS|}$ denotes the basis vector in $\R^{|\gS|}$ whose $s$-th element is one and others are zero, and $\vpi(s)\in\R^{|\gA|}$ is a column vector whose $\pi(s)$-th element is one and others are all zero. We will denote the greedy policy induced by $\mQ\in\R^{|\gS||\gA|}$, as $\pi_{\mQ}(s):=\arg\max_{a\in\gA}\mQ(s,a)$ where $\mQ(s,a)=(\ve_s\otimes\ve_a)^{\top} \mQ$, and let $\mPi^{\mQ}:=\mPi^{\pi_{\mQ}}$.

% \subsection{Synchronous model-based Q-learning}
% \import{prelim}{sync-q}

\subsection{Switched system theory}
In this section, a brief overview of the switching system~\citep{liberzon2005switched} is given. A switched system viewpoint has been a useful tool to analyze the behavior of $Q$-learning~\citep{lee2019unified}. The state variable $\vx_k\in\R^n, k \in\sN$ of a switched affine system evolves via the following equation:
\begin{align*}
    \vx_{k+1} = \mA_{\sigma(k)} \vx_k  + \vb_{\sigma(k)}, \quad \vx_0 \in\R^n
\end{align*}
where $\sigma\in\gM:\{1,2,\dots,|\gM| \}$ is called the mode, $\sigma(k)\in\gM$ is the switching signal, $\{\mA_{\sigma}\in\R^{n\times n };\sigma \in \gM\}$ and  $\{\vb_{\sigma} \in \R^n ; \sigma \in \gM\}$ are the subsystem matrices and vectors, respectively. The switching signal can be either determined arbitrary or controlled by a particular logic. When $\vb_{\sigma}$, the so-called affine term, is zero vector for any $\sigma\in\gM$, the above system is called the switched linear system.

\section{Synchronous Model-based Q-learning}

In this section, we first explicitly state the SyncMBQ in Algorithm~\ref{algo:sync_q} studied in this paper, and provide its sample complexity. We begin by introducing the estimators for the transition matrix and reward function.

\begin{algorithm}
	\caption{Synchronous Model-based Q-learning (SyncMBQ)} \label{algo:sync_q}
	\begin{algorithmic}[1]
        \State Initialize $\alpha \in (0,1],\; \mQ_1=\mQ_2=\dots=\mQ_m\in\R^{|\gS||\gA|},\;\hat{\mP}_{0}=\bm{0},\; \hat{\mR}_{0}=\bm{0}$.
			\For {$k=1,2,\dots,m$}
            \State Observe $s_k,a_k,s^{\prime}_k$ and $r_k:=r(s_k,a_k,s^{\prime}_k)$
             \State Update $\hat{\mP}_k$ and $\hat{\mR}_k$ following~(\ref{eq:update_hat_P_k}) and~(\ref{eq:update_hat_r_k}), respectively.
            \State Update $N_{k}^{s_ka_k}=N_{k-1}^{s_ka_k}+1$.
		\EndFor	
  \For {$k=m+1,m+2,\dots,$}
            \State Observe $s_k,a_k,s^{\prime}_k$ and $r_k:=r(s_k,a_k,s^{\prime}_k)$ where $s_k,a_k\sim d(\cdot)$ and $s_k^{\prime}\sim \gP(\cdot\mid s_k,a_k)$. %{\color{blue}[Here, add more details on how to sample the transition, i.e., $(s_k,a_k) \sim d$ and $s_k' \sim P(\cdot |s_k,a_k)$]}
             \State Update $\hat{\mP}_k$ and $\hat{\mR}_k$ following~(\ref{eq:update_hat_P_k}) and~(\ref{eq:update_hat_r_k}), respectively.
            \State Update $\mQ_k$:
            \begin{align}
                \mQ_{k+1} = \mQ_k + \alpha(\hat{\mR}_k+\gamma  \hat{\mP}_k \mPi_{\mQ_k} \mQ_k -  \mQ_k) . \label{eq:sync_q_update}
            \end{align}
        \State Update $N_{k}^{s_k,a_k}=N_{k-1}^{s_k,a_k}+1$.
		\EndFor
	\end{algorithmic} 
\end{algorithm}

\subsection{Model-based approach}

The model-based approach maintains statistical estimators of the transition probability $\gP(s,a,s^{\prime})$, denoted by $\hat{p}^{s,a,s^{\prime}}_{k}$, and expected reward $\E[\gR(s,a,s^{\prime})\mid s,a]$, denoted by $\hat{r}^{s,a}_k$, at time step $k\in\sN$ and for all $(s,a,s^{\prime})\in\gS\times\gA\times\gS$. We can naturally incorporate these estimators into the $Q$-learning update in~(\ref{eq:model-free-Q}), which can be written as follows:
\begin{align}
    \mQ_{k+1}(s_k,a_k)
    = & (1-\alpha) \mQ_k(s_k,a_k)                                                                                             \\
      & +\alpha \bigl(\hat{r}^{s_k,a_k}_k+\gamma \sum_{s\in\gS}\hat{p}^{s_k,a_k,s}_k\max_{a\in\gA}\mQ_k(s,a)\bigr). \nonumber
\end{align}

At time step $k\in\sN$, a simple choice for the estimators $\hat{p}^{s,a,s^{\prime}}_{k}$ and $\hat{r}^{s,a}_k$ is by a simple averaging rule over the past observations up to iteration $k\in\sN$:
\begin{align}
    \hat{p}^{s,a,s^{\prime}}_{k} = \frac{N^{s,a,s^{\prime}}_k}{N^{s,a}_k}, \quad \hat{r}^{s,a}_k= \frac{\sum_{i=0}^k r_i \bm{1}(\{ (s_i,a_i) = (s,a)  \})}{N^{s,a}_k}, \label{eq:moving_avg}
\end{align}
where $N^{s,a}_k$ denotes number of visits to state-action pair $(s,a)\in\gS\times \gA$ and $N^{s,a,s^{\prime}}_k$ is the number of visits to $(s,a,s^{\prime})\in\gS\times\gA\times\gS$ up to time $k\in\sN$, and $\bm{1}(A)$ is an indicator function such that returns one if event $A$ is true and otherwise zero.

Furthermore, with the estimation of the model, we can implement the update in a synchronous manner, i.e., for all $(s,a)\in \gS \times \gA$,
\begin{align}
    \mQ_{k+1}(s,a)
    = & (1-\alpha) \mQ_k(s,a)                                                                                                                    \\
      & +\alpha \bigl(\hat{r}^{s,a}_k+\gamma \sum_{s^{\prime}\in\gS}\hat{p}^{s,a,s^{\prime}}_k\max_{a\in\gA}\mQ_k(s^{\prime},a)\bigr), \nonumber
\end{align}
which corresponds to Algorithm~\ref{algo:sync_q}. Note that in the update~(\ref{eq:model-free-Q}), since we have only access to one tripe $(s_k,a_k,s_k^{\prime})$, we cannot update the iterate $\mQ_k$ for all $(s,a)\in\gS\times\gA$, which corresponds to the asynchronous update. Moreover, when $\alpha=1$, the above update corresponds to the online version of $Q$-value iteration.

The matrix notations corresponding to~(\ref{eq:moving_avg}) are defined as $\hat{\mP}_k\in \R^{|\gS||\gA|\times |\gS|}$ and $\hat{\mR}_k\in \R^{|\gS||\gA|}$, respectively. The $(s-1)|\gA|+a$-the row of $\hat{\mP}_k$ equals $\hat{\vp}_k^{s,a}\in\R^{|\gS|}$ whose $s^{\prime}$-th element is $\hat{p}^{sas^{\prime}}$. $\hat{\mR}_k$ is a vector whose $(s-1)|\gA|+a$-th element is $\hat{r}^{s,a}_k$. At time step $k$, upon observing $s_k,a_k,s_k^{\prime}$, we can use the following update rule for~(\ref{eq:moving_avg}):
\begin{align}
    \hat{\vp}_{k}^{s_k,a_k}=    & \frac{N^{s_k,a_k}_{k-1}}{N_{k-1}^{s_k,a_k}+1}\hat{\vp}_{k-1}^{s_k,a_k}+ \frac{1}{N_{k-1}^{s_k,a_k}+1}\ve_{s^{\prime}_k} ,\label{eq:update_hat_P_k} \\
    [\hat{\mR}_{k}]_{s_k,a_k} = & \frac{N^{s_ka_k}_{k-1}}{N^{s_ka_k}_{k-1}+1}[\mR_{k-1}]_{s_k,a_k}+ \frac{1}{N^{s_k,a_k}_{k-1}+1} r_{k} \label{eq:update_hat_r_k},
\end{align}

% \begin{align*}
%     \hat{\mP}_k := & \begin{bmatrix}
%                          \hat{\vp}^{1,1}_k &
%                          \hat{\vp}^{1,2}_k &
%                          \cdots            &
%                          \hat{\vp}^{|\gS|,|\gA|}_k
%                      \end{bmatrix}^{\top}  , \\
%     \hat{\mR}_k := & \begin{bmatrix}
%                          \hat{r}^{1,1}_k &
%                          \hat{r}^{1,2}_k &
%                          \cdots          &
%                          \hat{r}^{|\gS|,|\gA|}_k
%                      \end{bmatrix}^{\top}  ,
% \end{align*}
% where $\hat{\vp}^{s,a}_k$ for $(s,a)\in\gS\times \gA$, $s^{\prime}$-th element is defined as $[\hat{\vp}^{s,a}_k]_{s^{\prime}}:= \hat{p}^{sas^{\prime}}_k$ for $s^{\prime}\in\gS$. 

% For the first $m$-steps, we will only update the estimates for $\hat{\mP}_k$ and $\hat{\mR}_k$. 
where $[\vv]_{s,a}:=(\ve_s\otimes\ve_a)^{\top}\vv$ for $\vv\in\R^{|\gS||\gA|}$.
The overall steps including the model estimation steps are summarized in Algorithm~\ref{algo:sync_q}.
% We first observe $s_k,a_k$, and then update the estimates $\hat{p}_k^{s_ka_ks_k^{\prime}}$ and $\hat{r}^{s_ka_k}_k$ following~(\ref{eq:update_hat_P_k}) and~(\ref{eq:update_hat_r_k}), and then update $\mQ_k$ following~(\ref{eq:sync_q_update}).

% In this section, we explicitly state the synchronous model-based $Q$-learning algorithm in Algorithm~\ref{algo:sync_q} and provide its sample complexity.
% \subsubsection{Model-based approach}
%\subsection{SyncMBQ}

At this point, we make some remarks on the existing model-based RL methods. A line of researches~\citep{azar2012sample,gheshlaghi2013minimax,li2020breaking,agarwal2020model} focuses on model-based approaches, where parametric models are learned first from a generative model (or a sampling oracle), and then dynamic programming algorithms are performed using the estimated models--which we call offline methods. In contrast, we study a setting in which the model and the Q-function are learned jointly in an online manner--where only a single transition is observed each iteration. We consider a sampling scenario that a state-action pair is sampled by a fixed stationary state-action distribution under an exploratory behavior policy, and then the corresponding next state is sampled--which is weaker than the generative model assumption used in the prior works. Similar to our setting, the works in~\cite{kearns2002near,brafman2002r,strehl2005theoretical,szita2010model} studied model-based RL algorithms without relying on a generative model, where explorations and single trajectory-based samples are used. Although our sampling model is somewhat stronger than these trajectory-based settings, the proposed algorithm remains simple and can be interpreted as a direct extension of model-free Q-learning.

Now, we provide some details of Algorithm~\ref{algo:sync_q}, which mainly consists of the two stages:
\begin{enumerate}
    \item[1.] Data collection stage : For the first several updates, we only update the estimators $\hat{\mP}_k$ and $\hat{\mR}_k$. This guarantees that every state-action pair will be updated. Considering that we are interested in the $l_{\infty}$-norm error, we can expect that the $l_{\infty}$-norm error will decrease after such event. We will denote the number of iterations of the first stage as $m:=\frac{1}{d_{\min}} \ln \left( \frac{2|\gS||\gA|}{\delta} \right)$. Under uniform random sampling setting, i.e., $d_{\min}=\frac{1}{|\gS||\gA|}$, the first stage only requires $\gO(|\gS||\gA|)$ number of samples. That is, we require only one transition for each state-action pair.
    \item[2.] Learning stage: During this stage, we exploit the model to update the iterate and the model. We observe a sample $(s_k,a_k,s_k^{\prime})\in\gS\times\gA\times \gS$ from an i.i.d. distribution and update $\mQ_k$, $\hat{\mP}_k$, and $\hat{\mR}_k$.
\end{enumerate}

\begin{remark}\label{rem:online}
    Our purpose of data collection stage is different from that of existing model-based RL algorithms. $Q$-value iteration algorithms in in~\cite{gheshlaghi2013minimax,li2020breaking,agarwal2020model}, collect all the samples at once before updating the $Q$-function estimator. Likewise, batch $Q$-learning algorithms in~\citep{kearns1998finite,kalathil2021empirical,sidford2018near,wainwright2019variance} collect sufficient number of fresh samples each iteration before updating the $Q$-function estimator. It is meant to collect sufficient number of  samples to guarantee an accurate model before the update of $Q$-function estimator. In contrast, we only require one observation for each state-action pair to construct a stochastic matrix. Furthermore, we note that this stage can be omitted with initializing the transition matrix with an arbitrary stochastic matrix, and the error by the wrong initialization can be bounded by strong law of large number. However, for simplicity of the proof, we introduce the data collection stage to construct the correct empirical transition matrix.
\end{remark}

The following lemma guarantees that every state-action pair is visited at least one time after $m$-steps.

\begin{lemma}\label{lem:every_state_action_pair}
    For $m=\frac{1}{d_{\min} }\ln \frac{2|\gS||\gA|}{\delta}$, with probability at least $1-\frac{\delta}{2}$, every state-action pairs are visited, i.e.,
    \begin{align*}
        \sP\left[N^{s,a}_m \geq 1 , \; \forall (s,a) \in \gS\times\gA\right] \geq 1- \delta/2.
    \end{align*}
\end{lemma}
The proof is given in Appendix~\ref{app:lem:every_state_action_pair} in the full version. To proceed, we will denote the event that every state-action pair is visited as $\gE$:
\begin{align}
    \gE:=\{ N^{s,a}_m \geq 1, \forall (s,a) \in \gS\times\gA\}\label{event:every_visit}.
\end{align}

\subsection{Switched system perspective on SyncMBQ}
This section provides switched system perspective on SyncMBQ to study its non-asymptotic behavior. Let us first re-write the update of SyncMBQ in~(\ref{eq:sync_q_update}) with coordinate transform $\tilde{\mQ}_k:=\mQ_k-\mQ^*$, for $k\in\sN$ as follows:
\begin{align}
    \tilde{\mQ}_{k+1} = & \tilde{\mQ}_k+\alpha(\gamma\hat{\mP}_k\mPi^{\mQ_k}\tilde{\mQ}_k-\tilde{\mQ}_k)                                                \\
                        & + \alpha(\hat{\mR}_k+\gamma\hat{\mP}_k\mPi^{\mQ_k}\mQ^*-\mQ^*) \nonumber                                                      \\
    =                   & \mA_k^{\mQ_k}\tilde{\mQ}_k+\alpha \vw_k + \alpha \gamma \hat{\mP}_k(\mPi^{\mQ_k}-\mPi^{\mQ^*})\mQ^*, \label{eq:tildeQ_update}
\end{align}
where,
\begin{align}
    \mA^{\mQ}_k & := \mI+\alpha(\hat{\mP}_k\mPi^{\mQ}-\mI),  \nonumber                    \\
    \vw_k :=    & \hat{\mR}_k-\mR+\gamma(\hat{\mP}_k-\mP)\mPi^{\mQ^*}\mQ^*\label{eq:vw_k}
\end{align}
and $\vw_k$ corresponds to the stochastic noise term. Note that we have used the Bellman optimal equation $\mQ^*=\mR+\gamma\mP\mPi^{\mQ^*}\mQ^*$ in the definition of $\vw_k$. The affine term, $\hat{\mP}_k (\mPi^{\mQ_k}-\mPi^{\mQ^*})\mQ^*$ in~(\ref{eq:tildeQ_update}), causes significant challenge in the analysis of the algorithm. To avoid such difficulty, we will adopt the switched system perspective of $Q$-learning in~\cite{lee2019unified}. In particular, we can construct a sequence of iterates $\{  \mQ_k^U\}_{k\in\sN}$ and $\{ \mQ_k^L\}_{k\in\sN}$, which upper and lower bounds $\mQ_k$, respectively, i.e., $\mQ^U_k\geq \mQ_k \geq \mQ^L_k$ for all $k\in\sN$. Then, from the following relation
\begin{align}
    \left\|\mQ_k-\mQ^*\right\|_{\infty} \leq   \max\left\{ \left\| \mQ^U_k-\mQ^*\right\|_{\infty}, \left\| \mQ^L_k-\mQ^*\right\|_{\infty} \right\}, \label{ineq:QleqQ^U,Q^L}
\end{align}
we obtain desired error bound upon bounding the error of the upper comparison and lower comparison system. Letting $\tilde{\mQ}^U_k=\mQ^U_k-\mQ^*$, and $\tilde{\mQ}^L_k=\mQ^L_k-\mQ^*$, the following updates yields the upper and lower bounded iterates for $\tilde{\mQ}_k$, respectively:
\begin{align}
    \tilde{\mQ}^U_{k+1} = & \mA_k^{\mQ_k}\tilde{\mQ}^U_{k+1}+\alpha\vw_k ,  \quad \mQ^U_0=\mQ_0, \label{eq:upper_comparison_update} \\
    \tilde{\mQ}^L_{k+1} = & \mA_k^{\mQ^*}\tilde{\mQ}^L_{k+1}+\alpha\vw_k ,\quad \mQ^L_0=\mQ_0\label{eq:lower_comparison_update}.
\end{align}

The governing dynamics of $\tilde{\mQ}^U_k$ and $\tilde{\mQ}^L_k$ are switched linear system, which does not include any affine terms.

% The derivation of sample complexity to achieve $\left\|\mQ_k-\mQ^*\right\|_{\infty} \geq \eps$ for $\eps>0$,. We construct an upper and lower comparison system such that $\mQ^U_k \geq \mQ_k \geq \mQ^L_k$:
% \begin{align}
%     \mQ^U_{k+1}-\mQ^*= ( \mI+\alpha(\gamma\mP\mPi_{\mQ_k}-\mI))(\mQ^U_k-\mQ^*)+\alpha\vw_k \label{eq:upper_comparison_update} \\
%     \mQ^L_{k+1}-\mQ^*= ( \mI+\alpha(\gamma\mP\mPi_{\mQ^*}-\mI)) (\mQ^L_k-\mQ^*)+\alpha\vw_k \label{eq:lower_comparison_update},
% \end{align}
% where, 
% \begin{align}
%     \vw_k :=& \hat{\mR}_k + \gamma\hat{\mP}_k\mPi^{\mQ_k}\mQ_k-\mR-\gamma\mP\mPi^{\mQ_k}\mQ_k \label{eq:vw_k}.
% \end{align}

\begin{lemma}\label{lem:Q_l<Q<Q_u}
    Consider $\mQ^U_k$ and $\mQ^L_k$ in~(\ref{eq:upper_comparison_update}) and~(\ref{eq:lower_comparison_update}), respectively. Then, we have
    \begin{align*}
        \mQ^L_k \leq \mQ_k \leq \mQ^U_k.
    \end{align*}
\end{lemma}

The detailed derivations are given in Appendix~\ref{app:sec:switched_derivation}. Note that our construction of the upper and lower comparison systems differs from that of~\cite{lee2019unified} such that the noise term $\vw_k$ does not include the current iterate $\mQ_k$. This enables us to apply i.i.d. concentration bound on $\hat{\mP}_k\mPi^{\mQ^*}\mQ^*$, which will be clear in the subsequent subsection.

Note that under the event $\gE$,  for $k\geq m$, $\left\| \mA^{\mQ}_{k}\right\|_{\infty}\leq 1-(1-\gamma)\alpha$ for all $\mQ\in\R^{|\gS||\gA|}$.

\begin{lemma}\label{lem:A^Q_k-bound}
    Assume that the event $\gE$ holds. Then, $\left\|\mA^{\mQ}_k \right\|_{\infty} \leq 1-(1-\gamma)\alpha$ for $k\geq m$ and $\mQ\in\R^{|\gS||\gA|}$.
\end{lemma}
The proof is given in Appendix~\ref{app:lem:A^Q_k-bound} of the full version. Note that if $\vw_k$ is a zero vector for all $k\in\sN$, then from~(\ref{eq:upper_comparison_update}) and~(\ref{eq:lower_comparison_update}), we will have $\left\|\tilde{\mQ}^U_k\right\|_{\infty}$ and $\left\|\tilde{\mQ}^L_k\right\|_{\infty}$ converging to a zero vector at geometric rate of $\gO\left(\exp(-1-(1-\gamma)\alpha k)\right)$ from Lemma~\ref{lem:A^Q_k-bound}. However, $\vw_k$ is not a zero vector, and we will show that $\vw_k$ can be controlled by the concentration inequality given in the subsequent section.

\subsection{Concentration inequality for \texorpdfstring{$\vw_k$}{e}}
In this section, we provide a concentration inequality for $\vw_k$ in~(\ref{eq:vw_k}). The concentration inequality for $\vw_k$ will play an important role in deriving the sample complexity result. In particular, from the law of large numbers, we would expect $\hat{\mP}_k\to \mP$ and $\hat{\mR}_k\to\mR$ as the number of samples increases, yielding $\vw_k$ asymptotically converging to a zero vector. Concentration inequalities can characterize how many samples are required for achieving desired level of an accuracy. The following lemma describes a concentration inequality for $\hat{\mP}_k $ and $\hat{\mR}_k$:
\begin{lemma}\label{lem:inf-norm-bound-P_k-R_k}
    For $k\in\sN$, we have
    \begin{align*}
        \sP\left[\left\| (\hat{\mP}_k -\mP)\mPi^{\mQ^*}\mQ^*\right\|_{\infty} \geq \eps \right] \leq & \frac{3|\gS||\gA|}{\exp(kd_{\min}(1-\gamma)^2\eps^2/4)} , \\
        \sP\left[\left\|\hat{\mR}_k-\mR \right\|_{\infty} \geq \eps\right] \leq                      & 3|\gS||\gA| \exp(-kd_{\min}\eps^2/4),
    \end{align*}
    for $\eps^2 \in \left[0,\min\left\{3,\frac{3}{(1-\gamma)^2}\right\}\right]$.
\end{lemma}

% \begin{lemma}\label{lem:inf-norm-bound-P_k-R_k}
% For $k\in\sN$, we have
%      \begin{align*}
%         \sP\left[\left\|\hat{\mP}_k-\mP \right\|_{\infty} \geq \eps\right] \leq & |\gS||\gA|2^{|\gS|+2} \exp(-kd_{\min}\eps^2),\\
%         \sP\left[\left\|\hat{\mR}_k-\mR \right\|_{\infty} \geq \eps\right] \leq & 3|\gS||\gA| \exp(-kd_{s,a}\eps^2).
%     \end{align*}
%  for $\eps^2 \in [0,0.75]$. 
% \end{lemma}

The proof follows from applying standard concentration inequalities for i.i.d. random variables and is given in Appendix~\ref{app:sec:lem:inf-norm-bound-P_k-R_k} of the full version. Similar concentration inequalities have been also used in~\cite{azar2012sample}.

Now, applying the union bound to the above result yields concentration inequality for $\vw_k$.
\begin{lemma}\label{lem:w_concentration_bound}
    For $k\in\sN$, we have
    \begin{align*}
        \sP\left[\left\|\vw_k\right\|_{\infty}\geq \eps \right] \leq   6 |\gS||\gA| \exp(-kd_{\min}(1-\gamma)^2\eps^2/16),
    \end{align*}
    for $\eps^2 \in \left[0,\min\left\{12,\frac{12\gamma^2}{(1-\gamma)^2}\right\}  \right]$.
\end{lemma}

% \begin{lemma}\label{lem:w_concentration_bound}
%     For $k\in \sN$, we have
%     \begin{align*}
%         \sP\left[\left\| \vw_k \right\|_{\infty} \geq \eps \right] \leq |\gS||\gA|2^{|\gS|+3}\exp\left(-k d_{\min}\frac{ \eps^2(1-\gamma)^2}{4\gamma^2}\right),
%     \end{align*}
%     for $\eps^2\in\left[0, \min \left\{3,3 \frac{\gamma^2}{(1-\gamma)^2}\right\}\right]$.
% \end{lemma}

The proof is given in Appendix~\ref{app:lem:w_concentration_bound} of the full version.

\subsection{Sample complexity of Algorihtm~\ref{algo:sync_q}}
% !TeX root = ../main.tex
In this section, we provide a sample complexity to achieve $\eps$-accurate estimate of the optimal $Q$-function, i.e., $\left\|\mQ_k-\mQ^* \right\|_{\infty} \leq \eps$. Noting that the upper and lower comparison systems in~(\ref{eq:upper_comparison_update}) and~(\ref{eq:lower_comparison_update}), respectively, share the same noise term $\vw_k$, both systems can be viewed as particular case of the following recursion:
\begin{align}
    \vx_{k+1} = \mA^{\vy_k}_k\vx_k + \alpha \vw_k, \quad \vx_0 \in \R^{|\gS||\gA|}, \label{eq:general_recursion}
\end{align}
where $\{\vy_k \in \R^{|\gS||\gA|}\}_{k\in\sN}$ is arbitrary sequence of vectors. If $\vx_0=\tilde{\mQ}^U_0$ and $\vy_k=\mQ_k, \;k\in\sN$, then $\vx_k$ coincides with $\tilde{\mQ}^U_k$ for $k\in\sN$. Likewise, when $\vx_0=\tilde{\mQ}^L_0$ and $\vy_k=\mQ^*$, then $\vx_k$ coincides with $\tilde{\mQ}^L_k$ for $k\in\sN$. Therefore, it suffices to bound $\left\|\vx_k \right\|_{\infty}$ from the relation~(\ref{ineq:QleqQ^U,Q^L}). Recursively expanding~(\ref{eq:general_recursion}) and applying concentration inequality on $\vw_k$ from Lemma~\ref{lem:w_concentration_bound} would yield the following result:

\begin{theorem}\label{thm:sample_complexity_proof}
    For $k\in \sN$, with probability at least $1-\delta$, we have $\left\|\mQ_k-\mQ^*\right\|_{\infty} \leq \eps$, with at most following number of samples:
    \begin{align*}
        \tilde{\gO}\left( \frac{1}{(1-\gamma)^4\eps^2d_{\min}} + \frac{1}{(1-\gamma)\alpha} \right),
    \end{align*}
    for $\eps^2 \in \left[0,\frac{36}{(1-\gamma)^2} \min \left\{12,3 \frac{\gamma^2}{(1-\gamma)^2}\right\} \right]$ and $\delta\in (0,1)$.
\end{theorem}

The proof is given in Appendix~\ref{app:thm:sample_complexity_proof} in the full version. As in the literature of non-asymptotic analysis of RL algorithms, our sample complexity bound depends on the so-called effective horizon $\frac{1}{1-\gamma}$ and minimum value of the sampling distribution $d_{\min}$. We improve over the sample complexity result of~\cite{lee2022finite}, which also relied on switched system analysis and provided $\tilde{\gO}\left( \frac{(|\gS||\gA|)^4}{(1-\gamma)^6\eps^2} \right)$ under the step-size $\alpha \in (0,1)$. Our bound implies sample complexity $\tilde{\gO}\left(  \frac{|\gS||\gA|}{(1-\gamma)^4\eps^2}\right)$ (with $d_{\min}= 1/(|{\cal S}||{\cal A}|)$). Lastly, we note that the restriction on $\epsilon$ is used for technical simplicity, and the bound is
larger than the maximum value of $Q$-functions, and given the boundedness of the iterate $\mQ_k$,
the requirement on $\epsilon$ can be removed when $\gamma$ is close to one.

Now, we compare our results with existing sample complexity guarantees for model-based RL algorithms. The works of \cite{azar2012sample,agarwal2020model} established the sample complexity bound $\tilde{\gO}\!\left(\frac{|\gS||\gA|}{(1-\gamma)^3\eps^2}\right)$, which matches the minimax lower bound for general RL problems. In particular, $Q$-value iteration is shown to achieve this optimal rate. However, these results are derived under offline learning setting, and the assumption of a generative model (or sampling oracle), where the learner can query any state-action pair and obtain independent samples of the next state according to the transition dynamics. Such algorithms therefore operate in an offline setting and rely on a stronger sampling model than the one considered in this paper.

In the online setting with single-trajectory data, the sample complexity of model-based RL algorithms has been studied in \cite{kearns2002near,brafman2002r,strehl2005theoretical,szita2010model,lattimore2014near}. Among these works, the sharpest guarantees were established in \cite{szita2010model} and \cite{lattimore2014near}, which obtained sample complexity bounds for computing an $\eps$-optimal policy of order $
    \tilde{\gO}\!\left(\frac{|\gS||\gA|}{(1-\gamma)^6\eps^2}\right)
    \quad\text{and}\quad
    \tilde{\gO}\!\left(\frac{|\gS|^2|\gA|}{(1-\gamma)^3\eps^2}\right)$, respectively. These approaches operate under a weaker sampling model than the i.i.d. sampling setting, as data are collected along a single trajectory induced by the learning policy. Nevertheless, their guarantees are either weaker than or comparable to the bound established in Theorem~\ref{thm:sample_complexity_proof}. In particular, an $\eps$-optimal policy can be obtained from an $\eps$-accurate $Q$-function estimate with an additional factor of $\frac{1}{1-\gamma}$ via the performance difference lemma \cite{kakade2003sample}. A summary of the overall comparisons is provided in Table~\ref{tab:sample_complexity}.

% Besides, it is also worth noting model-based RL algorithms that in online and follows a single trajectory~\citep{kearns2002near,brafman2002r,szita2010model,strehl2005theoretical,lattimore2014near}. In particular,~\cite{szita2010model} and~\cite{lattimore2014near} obtained the sample complexities to compute the $\eps$-optimal policy, $\tilde{\gO}\left(\frac{|\gS||\gA|}{(1-\gamma)^6\eps^2}\right)$ and $\tilde{\gO}\left(\frac{|\gS|^2|\gA|}{(1-\gamma)^3\eps^2}\right)$, respectively. Even though their algorithms do not rely on a generative model similar to ours and use slightly different sample complexity measures, they exhibit worse or comparable sample complexities than ours in Theorem~\ref{thm:sample_complexity_proof}.

%Moreover, as mentioned in Remark~\ref{rem:online}, it is not an online algorithm, where a single sample is observed every iteration.~\cite{wainwright2019variance} applied a variance-reduction technique~\citep{johnson2013accelerating} to the $Q$-learning with diminishing step-size, and proved that it can achieve sample complexity $\tilde{\gO}\left(\frac{|\gS||\gA|}{(1-\gamma)^3\eps^2}\right)$, which is the optimal sample complexity of RL algorithms. The variance reduction technique can also be considered as implicitly estimating the model because the parameter, that the variance reduction technique tries to approximate, includes the transition matrix and reward function. As previously mentioned,~\cite{wainwright2019variance} relies on generative model and uses diminishing step-size, whereas we focus on online sampling setting and constant step-size.

\begin{table}[t]
    \centering
    \footnotesize
    \begin{tabular}{lccc}
        \toprule
        \textbf{Name}                 & \textbf{Sample Complexity}                                                       & \textbf{Sampling Model} & \textbf{Criteria} \\
        \midrule
        ~\cite{azar2012sample}        & $\mathcal{O}\!\left(\frac{1}{(1-\gamma)^4\varepsilon^2}\right)$                  & generative/offline      & value function    \\

        ~\cite{gheshlaghi2013minimax} & $\mathcal{O}\!\left(\frac{|\gS||\gA|}{(1-\gamma)^4\varepsilon^2}\right)$         & generative/offline      & policy            \\

        ~\cite{li2020breaking}        & $\tilde{\mathcal{O}}\!\left(\frac{|\gS||\gA|}{(1-\gamma)^3\varepsilon^2}\right)$ & generative/offline      & policy            \\
        ~\cite{agarwal2020model}      & $\tilde{\mathcal{O}}\!\left(\frac{|\gS||\gA|}{(1-\gamma)^3\varepsilon^2}\right)$ & generative/offline      & policy            \\
        ~\cite{kearns2002near}        & $\tilde{\gO}\left( \frac{|\gS|^2|\gA|}{(1-\gamma)^6\eps^3} \right)$              & episodic/online         & policy            \\
        ~\cite{brafman2002r}          & $\tilde{\gO}\left( \frac{|\gS|^2|\gA|}{(1-\gamma)^6\eps^3} \right)$              & episodic/online         & policy            \\
        ~\cite{strehl2005theoretical} & $\tilde{\gO}\left( \frac{|\gS|^2|\gA|}{(1-\gamma)^6\eps^3} \right)$              & episodic/online         & policy            \\
        ~\cite{szita2010model}        & $\tilde{\gO}\left(\frac{|\gS||\gA|}{(1-\gamma)^6\eps^2}\right)$                  & episodic/online         & policy            \\
        ~\cite{lattimore2014near}     & $\tilde{\gO}\left( \frac{|\gS|^2|\gA|}{(1-\gamma)^3\eps^2} \right)$              & episodic/online         & policy            \\
        Ours                          & $\tilde{\gO}\left( \frac{|\gS||\gA|}{(1-\gamma)^4\eps^2}  \right)$               & i.i.d  /online          & value function    \\

        \bottomrule
    \end{tabular}
    \caption{Comparison of Model-based RL Algorithms. One can convert the $\eps$-optimal valeu function criteri to $\eps$-optimal policy criteria with additional factor $\frac{1}{1-\gamma}$ using the performance difference lemma~\citep{kakade2003sample}.}
    \label{tab:sample_complexity}
\end{table}

% Noting that $ \left\| \mA_{\mQ} \right\|_{\infty} \leq 1-(1-\gamma)\alpha$ for any $\mQ\in\R^{|\gS||\gA|}$, and $\vw_k$ is same in the update in~(\ref{eq:upper_comparison_update}) and~(\ref{eq:lower_comparison_update}), it is enough to bound the following type of recursion :
% \begin{align}
%     \vx_{k+1}= \mA_{\vy_k}\vx_k+\vw_k.\label{eq:general_sa_with_wk}
% \end{align}
% where $\{\vy_i\in \R^{|\gS||\gA|}\}_{i=1}^k$ is sequence of arbitrary vectors. 

% \begin{proposition}\label{prop:sample_complexity}
%     With probability at least $1-\delta$, $\left\|\vx_k\right\|_{\infty} \leq \eps$ , with at most following number of samples:
%     \begin{align*} 
%            \tilde{\gO}\left( \frac{|\gS|}{(1-\gamma)^4\eps^2d_{\min}} + \frac{1}{(1-\gamma)\alpha} \right),
%     \end{align*}
%     for $\eps^2 \in \left[0,\frac{36}{(1-\gamma)^2} \min \left\{3,3 \frac{\gamma^2}{(1-\gamma)^2}\right\} \right]$.
% \end{proposition}

% The proof is given in Appendix Section~\ref{app:prop:sample_complexity}.

% \begin{theorem}
%     For $k\in \sN$, with probability at least $1-\delta$, we have $\left\|\mQ_k-\mQ^*\right\|_{\infty} \leq \eps$, with at most following number of samples:
%         \begin{align*} 
%            \tilde{\gO}\left( \frac{|\gS|}{(1-\gamma)^4\eps^2d_{\min}} + \frac{1}{(1-\gamma)\alpha} \right),
%     \end{align*}
%     for $\eps^2 \in \left[0,\frac{36}{(1-\gamma)^2} \min \left\{3,3 \frac{\gamma^2}{(1-\gamma)^2}\right\} \right]$.
% \end{theorem}

\section{Experiments}
\newcommand{\cmss}[1]{{\fontfamily{cmss}\selectfont{#1}}}

\definecolor{AlloyOrange}{RGB}{200, 86, 20}
\definecolor{CyanCornflowerBlue}{RGB}{20, 134, 200}

In this section, we first empirically show the correctness of suggested error bound in a simple MDP. Then, we investigate performance in two benchmark environments. In the experiments, we used the discount factor $\gamma=0.9$; $\epsilon$-greedy behavior with $\epsilon=0.1$; and tabular action-values initialized with $0$.

\subsection{Error bound}
We firstly visualize the evolution of error $\left\|\mQ_k -\mQ^*\right\|_{\infty}$ according to $k$. According to Theorem~\ref{thm:sample_complexity_proof}, it is ideal for the value to decrease proportional to $k$.

\begin{figure}
    \centering
    \includegraphics[width=0.5\linewidth]{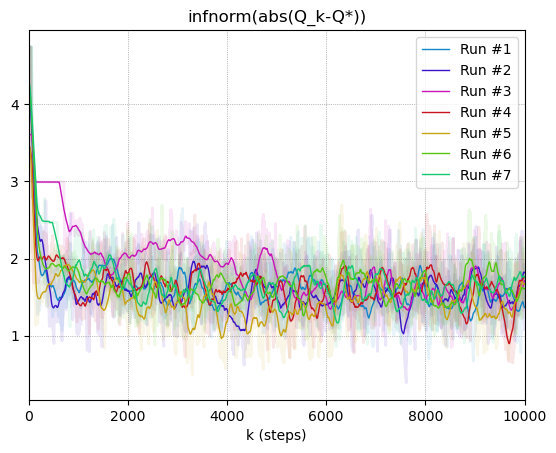}
    \caption{Showing decreasing error between $\mQ_k$ and $\mQ^*$ in a random MDP. Seven runs for the same MDP are conducted. Moving averages are highlighted as vivid line.}
    \label{fig:error-bound}
\end{figure}

We artificially construct a simple stochastic MDP with $|\gS|=4$ and $|\gA|=4$.
One of the states is set to the terminal state, and one of the rest is set to the starting state. The transition probability and reward functions are randomly generated. After constructing an MDP, we tested seven runs under constant step-size $\alpha=0.1$. In Fig~\ref{fig:error-bound}, every state-action pair is visited after about 65-th step. The error dramatically decreased at first, then keeps slowly decreasing.
Although the error still has deviation from 0 which means $\mQ_k = \mQ^*$, we manually checked that $\pi_{\mQ_k} = \pi^*$ sufficiently holds.

\subsection{Performance on Benchmark Environments}

\begin{figure*}[!ht]
    \centering
    \begin{subfigure}[c]{0.28\linewidth}
        \centering
        \includegraphics[width=\linewidth]{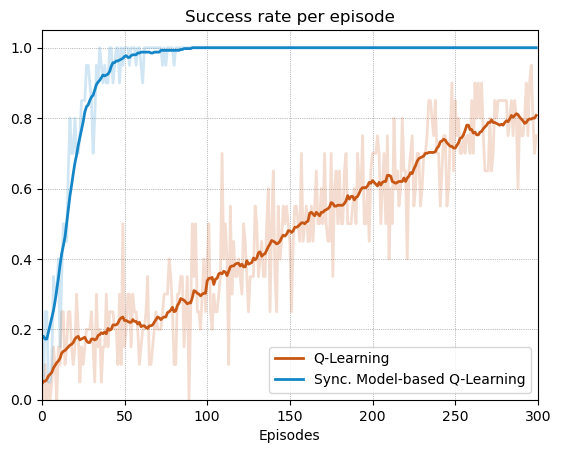}
        \caption{\cmss{Taxi}, $\alpha=0.1$}
    \end{subfigure}
    \hfill
    \begin{subfigure}[c]{0.28\linewidth}
        \centering
        \includegraphics[width=\linewidth]{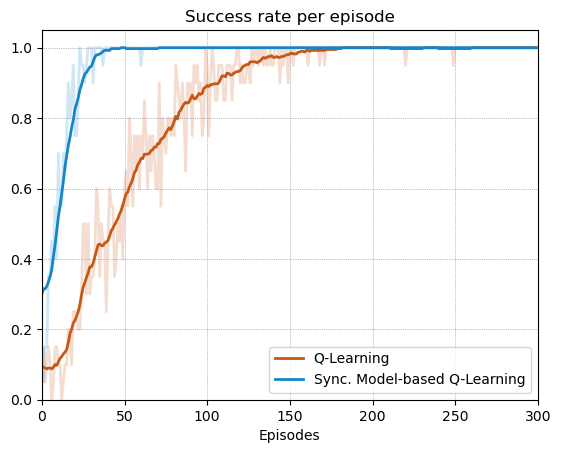}
        \caption{\cmss{Taxi}, $\alpha=0.5$}
    \end{subfigure}
    \hfill
    \begin{subfigure}[c]{0.28\linewidth}
        \centering
        \includegraphics[width=\linewidth]{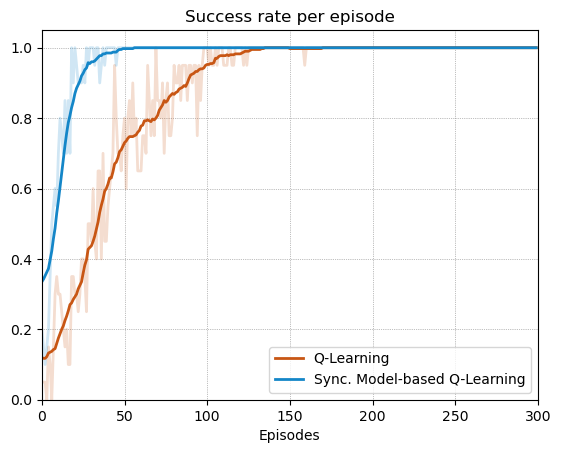}
        \caption{\cmss{Taxi}, $\alpha=0.9$}
    \end{subfigure}
    \hfill
    \begin{subfigure}[c]{0.28\linewidth}
        \centering
        \includegraphics[width=\linewidth]{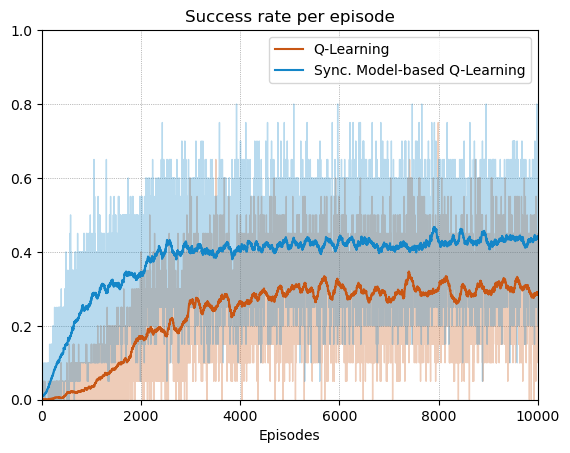}
        \caption{\cmss{FrozenLake}, $\alpha=0.1$}
    \end{subfigure}
    \hfill
    \begin{subfigure}[c]{0.28\linewidth}
        \centering
        \includegraphics[width=\linewidth]{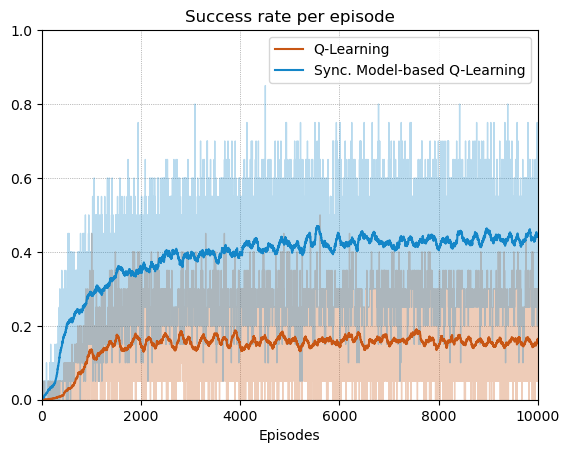}
        \caption{\cmss{FrozenLake}, $\alpha=0.5$}
    \end{subfigure}
    \hfill
    \begin{subfigure}[c]{0.28\linewidth}
        \centering
        \includegraphics[width=\linewidth]{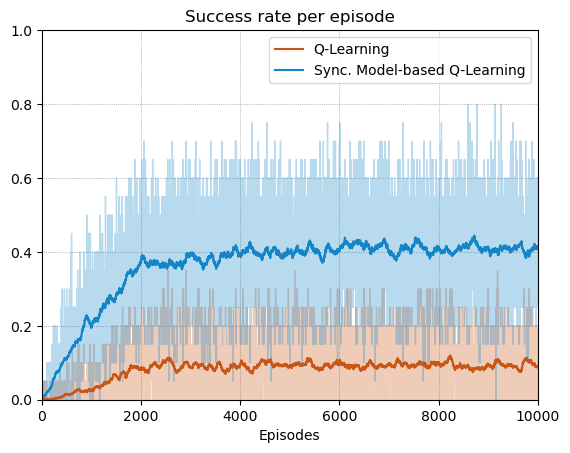}
        \caption{\cmss{FrozenLake}, $\alpha=0.9$}
    \end{subfigure}
    \hfill
    \vfill
    \begin{subfigure}[c]{0.4\linewidth}
        \centering
        \begin{tabular}{c|cc}
            $\alpha$ & \textcolor{AlloyOrange}{Q}  & \textcolor{CyanCornflowerBlue}{SyncMBQ} \\
            \midrule
            0.1      & 0.95\scriptsize{$\pm$1.20}  & 98.30\scriptsize{$\pm$4.83}             \\
            0.5      & 45.35\scriptsize{$\pm$5.99} & 98.75\scriptsize{$\pm$3.24}             \\
            0.9      & 61.00\scriptsize{$\pm$3.95} & 88.25\scriptsize{$\pm$9.80}             \\
        \end{tabular}
        \caption{\cmss{Taxi}, Success rate (\%) under greedy policy after training}
    \end{subfigure}
    \hfill
    \begin{subfigure}[c]{0.4\linewidth}
        \centering
        \begin{tabular}{c|cc}
            $\alpha$ & \textcolor{AlloyOrange}{Q}   & \textcolor{CyanCornflowerBlue}{SyncMBQ} \\
            \midrule
            $0.1$    & 47.15\scriptsize{$\pm$19.00} & 71.25\scriptsize{$\pm$4.44}             \\
            $0.5$    & 19.15\scriptsize{$\pm$15.22} & 71.10\scriptsize{$\pm$4.85}             \\
            $0.9$    & 18.35\scriptsize{$\pm$15.01} & 54.20\scriptsize{$\pm$16.66}            \\
        \end{tabular}
        \caption{\cmss{FrozenLake}, Success rate (\%) under greedy policy after training}
    \end{subfigure}

    \caption{Perfomance of the synchronous model-based $Q$-learning. \cmss{Taxi} (top row) and \cmss{FrozenLake} (bottom row). For graphs, moving averages are highlighted as vivid line with a window size of 20 episodes for \cmss{Taxi} and 100 for \cmss{FrozenLake}. For tables, the mean and standard deviation averaged over 20 runs are shown.}
    \label{fig:exp-gym}
\end{figure*}

To evaluate the proposed  SyncMBQ, we choose two environments from Gymnasium~\citep{towers_gymnasium_2023}: \cmss{Taxi-v3} and \cmss{FrozenLake8x8-v1}. Experiments were conducted using constant step-size of $\alpha \in \{0.1, 0.5, 0.9\}$ and the results are averaged over 20 runs for each algorithm.

\cmss{Taxi} is a deterministic environment with $|\gS|=500$ and $|\gA|=6$. Each episode starts at random one of 300 possible starting states. Reward of $-1$ is given for each step, unless for a wrong pick-drop ($-10$) or for a successful drop ($+20$). An episode terminates after the drop action.
\cmss{FrozenLake} is a stochastic environment with $|\gS|=64$ and $|\gA|=4$. The agent will move in intended direction with probability of $1/3$ else will move in either perpendicular direction with equal probability of $1/3$ in both directions. All episodes start at the top left starting point. The only reward of $1$ is given when the agent reaches the goal. An episode terminates when the agent fells into a hole or reaches the goal.

We see from Figure \ref{fig:exp-gym} that SyncMBQ performs better than the original $Q$-learning, especially in \cmss{FrozenLake} which is an stochastic environment.
The left three columns show the average success rate while training. In \cmss{Taxi} shown at top row, SyncMBQ achieves full success on its early stage, while $Q$-learning gradually converges to the optimal policy. Similarly in \cmss{FrozenLake} at bottom row, the success rate of SyncMBQ is always higher then the standard $Q$-learning.
We can get another advantage with SyncMBQ. While the performance of $Q$-learning depends a lot on learning rate $\alpha$, SyncMBQ shows stable outperformance.
The rightmost column is about the success rate of greedy policy, since $\epsilon$-greedy behavior policy is used until the end of the training. This quantitative results again proves the superiority of SyncMBQ.

\section{Conclusion}
In this paper, we have studied the non-asymptotic behavior and empirical performance of SyncMBQ, which is direct extension of $Q$-learning to the model-based setting. We have considered online learning setting and relaxed sampling model which is weaker assumption than the generative model assumption. Furthermore, we proved that SyncMBQ can achieve the optimal sample complexity of $Q$-learning with general step-size. Moreover, we developed a new switched system model for $Q$-learning, and experimentally verified the superiority of SyncMBQ over $Q$-learning. Future studies include considering the setting where we collect samples following a single trajectory and dealing with the exploration-exploitation dilemma.

%%%%%%%%%%%%%%%%%%%%%%%%%%%%%%%%%%%%%%%%%%%%%%%%%%%%%%%%%%%%%%%%%%%%%%%%%%%%%%%%
\section{ACKNOWLEDGMENTS}

The work was supported by the Institute of Information
Communications Technology Planning Evaluation (IITP)
funded by the Korea government under Grant 2022-0-00469,
and the BK21 FOUR from the Ministry of Education (Republic of Korea).

\bibliographystyle{IEEEtran}
{\small\sloppy\bibliography{biblio}\par}

\appendices

\section{Notations}
$\sN$ : set of natural numbers; $\R^n,\;n\in\sN$ : set of real-valued $n$-dimensional vectors; $\R^{n\times m},\;n.m\in\sN$ : set of real-valued $n\times m$-dimensional matrices; $[\vv]_i$ for $\vv\in\R^n$ and $1\leq i \leq n$: $i$-th element of $\vv$; $[\mA]_{i,j}$ for $\mA\in \R^{n\times m},\;1\leq i \leq n, \; 1\leq j \leq m$: $i$-th row and $j$-th column element of $\mA$; $\otimes$ : Kronecker product; $\lceil a \rceil $ for $a\in\R$ : the smallest integer greater than or equal to $a\in\R$; $\lfloor a \rfloor $ for $a\in\R$ : the greatest integer less than or equal to $a\in\R$.

\section{Upper and lower comparison system derivation}\label{app:sec:switched_derivation}

% We will first derive the upper comparison system. We can re-write~(\ref{eq:syncq_coordinate_change}) as follows:
% \begin{align*}
%     \tilde{\mQ}_{k+1} =& \tilde{\mQ}_k + \alpha (\mR+\gamma\mP\mPi^{\mQ_k}\tilde{\mQ}_k -\tilde{\mQ}_k +\gamma\mP\mPi^{\mQ_k}\mQ^*-\mQ^*)+\alpha \vw_k \\
%       =& (\mI+\alpha (\gamma \mP\mPi^{\mQ_k}-\mI)\tilde{\mQ}_k + \gamma \mP(\mPi^{\mQ_k}-\mPi^{\mQ^*})\mQ^*+\alpha\vw_k.
% \end{align*}
\begin{lemma}\label{lem:Q<Q_u}
    For $k\in\sN$, we have $\mQ^U_k \geq \mQ_k $,
\end{lemma}
\begin{proof}
    The proof follows from induction on the hypothesis $\mQ^U_k \geq \mQ_k$. Suppose that the argument holds for some $k\in\sN$. Then, letting $\tilde{\mQ}^U_k=\mQ^U_k-\mQ^*$, we have $\tilde{\mQ}^U_k\geq \tilde{\mQ}_k$. Now, we show that the following argument holds for $\tilde{\mQ}^U_{k+1}$:
    \begin{align*}
        \tilde{\mQ}^U_{k+1} = & \mA^{\mQ_k}_k\tilde{\mQ}^U_k + \alpha \vw_k                                                       \\
        \geq                  & \mA^{\mQ_k}_k \tilde{\mQ}_k+ \alpha\gamma \hat{\mP}_k(\mPi^{\mQ_k}-\mPi^{\mQ^*})\mQ^*+\alpha\vw_k \\
        =                     & \tilde{\mQ}_{k+1},
    \end{align*}
    where the first inequality follows from the fact that $\mA^{\mQ_k}_k$ is a positive matrix, i.e., the elements of $\mA^{\mQ_k}_k$ are all non-negative, and the fact that $\mPi^{\mQ^*}\mQ^*\geq \mPi^{\mQ_k}\mQ^*$. The proof is completed by the induction argument.
\end{proof}

\begin{lemma}\label{lem:Q_l<Q}
    For $k\in\sN$, we have $\mQ^L_k \leq \mQ_k$.
\end{lemma}

\begin{proof}
    Let $\tilde{\mQ}^L_k:=\mQ^L_k-\mQ^*$. Suppose the argument holds for some $k\in\sN$. Then, we have
    \begin{align*}
        \tilde{\mQ}^L_{k+1} = & \mA^{\mQ^*}_k \tilde{\mQ}^L_k +\alpha\vw_k                                                                                  \\
        =                     & (1-\alpha)\tilde{\mQ}^L_k+ \alpha\gamma\hat{\mP}_k\mPi^{\mQ^*}\tilde{\mQ}^L_k + \alpha \vw_k                                \\
        \leq                  & (1-\alpha)\tilde{\mQ}_k + \alpha\gamma\hat{\mP}_k\mPi^{\mQ^*}\mQ^L_k -\alpha\gamma\hat{\mP}_k\mPi^{\mQ^*}\mQ^*+\alpha \vw_k \\
        \leq                  & (1-\alpha)\tilde{\mQ}_k + \alpha\gamma\hat{\mP}_k\mPi^{\mQ_k}\mQ_k -\alpha\gamma\hat{\mP}_k\mPi^{\mQ^*}\mQ^*+\alpha\vw_k    \\
        =                     & ((1-\alpha)\mI+\gamma\hat{\mP}_k\mPi^{\mQ_k})\tilde{\mQ}_k +\alpha\gamma \hat{\mP}_k\mPi^{\mQ_k}\mQ^*                       \\
                              & -\alpha\gamma\hat{\mP}_k\mPi^{\mQ^*}\mQ^*+\alpha\vw_k                                                                       \\
        =                     & \tilde{\mQ}_{k+1},
    \end{align*}
    where the first inequality follows from the induction hypothesis, the second inequality follows from the relation $\mPi^{\mQ^*}\mQ^L_k\leq \mPi^{\mQ^L_k}\mQ^L_k\leq \mPi^{\mQ^L_k}\mQ_k\leq \mPi^{\mQ_k}\mQ_k $, and the last equality follows from~(\ref{eq:tildeQ_update}). The proof is completed by the induction argument.
\end{proof}

\subsection{Proof of Lemma~\ref{lem:Q_l<Q<Q_u}}

\begin{proof}
    The proof is readily completed using Lemma~\ref{lem:Q_l<Q} and~\ref{lem:Q<Q_u}.
\end{proof}

\section{Technical lemmas}\label{app:sec:technical}
\subsection{Concentration inequalities}

\begin{lemma}[Theorem B.6 in~\cite{shalev2014understanding}]\label{lem:binom_tail_bound}
    Let $X_1,\dots, X_n$ be independent random variables such that $\E[X_i]=\mu$ and $  a\leq X_i \leq b$ for every $i\in \sN$. Then for any $\eps > 0$, we have
    \begin{align*}
        \sP \left[\left|\frac{1}{n}\sum_{i=1}^n (X_i - \mu  ) \right| \geq \eps \right] \leq 2 \exp ( -2 n \eps^2/(b-a)^2 ).
    \end{align*}
\end{lemma}

% \begin{lemma}[Concentration inequality for $\left\|\cdot\right\|_1$]\label{lem:l1-norm-conentration}
%     Consider a sequence of i.i.d. random vectors $\{\vv_i\in\R^d\}_{i=1}^n$ such that  $\E[\vv_i]=\vv$ and $\left\|\vv_i\right\|_1 \leq 1$ for $1\leq j \leq d$.  Then, we have
%     \begin{align*}
%         \sP\left[\left\| \frac{1}{n}\sum_{i=1}^n \vv_i -\vv \right\|_1 \geq \eps \right] \leq 2^{d+1} \exp(-n\eps^2/2).
%     \end{align*}
% \end{lemma}
% \begin{proof}
%     We have for $\vx\in\R^d$, $\left\|\vx\right\|_1 = \max_{\vu\in\{-1,1\}^d} \vu^{\top}\vx$. Therefore, 
%     \begin{align*}
%         \sP \left[ \left\| \frac{1}{n}\sum_{i=1}^n \vv_i -\vv \right\|_1 \geq \eps\right]  =& \sP\left[ \max_{\vu\in\{-1,1\}^{d}} \vu^{\top}\left(\frac{1}{n}\sum^n_{i=1}\vv_i-\vv\right)\geq \eps \right] \\
%         \leq & \sum_{\vu\in\{-1,1\}^{d}} \sP\left[ \vu^{\top}\left( \frac{1}{n}\sum^n_{i=1} \vv_i-\vv\right) \geq \eps \right] \\
%         \leq & 2^{d+1} \exp(-2n\eps^2).
%         \end{align*}
%          The first inequality follows from union bound. Noting that for any $\vu\in \{-1,1\}^d$ and $1\leq i\leq n$, $\E[\vu^{\top}\vv_i]=\vu^{\top}\E[\vv_i]=\vu^{\top}\vv$, and $|\vu^{\top}\vv_i|\leq \left\|\vu\right\|_{\infty}\left\|\vv_i\right\|_1 \leq 1 $, the second inequality follows from Lemma~\ref{lem:binom_tail_bound} in the Appendix Section~\ref{app:sec:technical}.
% \end{proof}

\begin{lemma}\label{lem:omega_bound}
    For $s\in\gS,a\in\gA,k\in\sN$, suppose that an event $\Omega$ depends on $s,a,k,\eps$, and state-action trajectory $\{(s_k,a_k,s_k^{\prime})\}_{k=1}^{\infty}$. Moreover, assume that for some positive constant $A,B$, the following holds:
    \begin{align}
        \sP\left[\Omega(s,a,k,\eps)\middle | N^{s,a}_k=t \right] \leq A \exp(- t B \eps^2) ,\label{ineq:omega_bound_condition}
    \end{align} for $1\leq t \leq k$, and $N^{s,a}_k$ denotes number of visits to state-action pair $(s,a)\in\gS\times\gA$. Then, we have
    \begin{align*}
        \sP\left[\Omega(s,a,k,\eps) \right] \leq A\exp(-kd(s,a)B\eps^2/2) + \exp(-kd(s,a)),
    \end{align*}
    for $\eps^2 \in [0,1.59/B]$.
\end{lemma}
\begin{proof}
    The law of total probability yields
    \begin{align}
             & \sP\left[ \Omega(s,a,k,\eps) \cap \{ N^{s,a}_k \geq 1 \}\right]                                     \\
        =    & \sum^{k}_{t=1} \sP\left[ \Omega(s,a,k,\eps) \middle | N^{sa}_k =t\right] \sP[N^{s,a}_k=t] \nonumber \\
        \leq & A \sum^k_{t=1}  \exp(-tB\eps^2) \binom{k}{t}d(s,a)^t(1-d(s,a))^{k-t} \nonumber                      \\
        =    & \sum^k_{t=1} A(d(s,a) \binom{k}{t}\exp(-B\eps^2))^t(1-d(s,a))^{k-t} \nonumber                       \\
        \leq & A( 1-d(s,a)+d(s,a)\exp(-B\eps^2))^k \nonumber                                                       \\
        \leq & A \left (1-d(s,a)\frac{B\eps^2}{2}\right)^k \nonumber                                               \\
        \leq & A \exp(-kd(s,a)B\eps^2/2), \label{ineq:omega_bound_1}
    \end{align}
    where the first inequality follows from the assumption, the second inequality follows from the binomial theorem, and the last inequality follows from $\exp(-x)\leq 1-\frac{x}{2}$ for $x\in[0,1.59]$. Next, it follows that
    \begin{align*}
        \sP\left[ \Omega (s,a,k,\eps) \right] = & \sP\left[\Omega (s,a,k,\eps)  \cap \{N^{sa}_k \geq 1 \}\right] \\
                                                & +\sP\left[\Omega (s,a,k,\eps)  \cap \{N^{s,a}_k =0  \right]    \\
        \leq                                    & A\exp(-kd(s,a)B\eps^2/2) + \sP[N^{s,a}_k=0]                    \\
        \leq                                    & A\exp(-kd(s,a)B\eps^2/2) +  (1-d(s,a))^k                       \\
        \leq                                    & A\exp(-kd(s,a)B\eps^2/2) + \exp(-kd(s,a)),
    \end{align*}
    where the first inequality follows from~(\ref{ineq:omega_bound_1}), the second inequality follows from the fact that $\sP[(s_t,a_t)=(s,a)]=d(s,a)$ for $1\leq t \leq k$, and the last inequality follows from the relation $1-x\leq\exp(-x)$ for $x\geq 0$.
\end{proof}

\subsection{Spectral properties}\label{app:property_Q}

\begin{lemma}\label{lem:Q_k_bound}
    For $k\in\sN$, we have  \(\left\| \mQ_k \right\|_{\infty} \leq \frac{1}{1-\gamma}\). Moreover, $\left\| \mQ^* \right\|_{\infty} \leq \frac{1}{1-\gamma}$.
\end{lemma}
\begin{proof}
    Suppose that the statement holds for some $k\in \sN$, i.e., $\left\|\mQ_k\right\|_{\infty}\leq \frac{1}{1-\gamma}$. Then, we have
    \begin{align*}
        \left\|\mQ_{k+1}\right\|_{\infty} \leq & (1-\alpha)\left\|\mQ_k\right\|_{\infty} + \alpha (1+\gamma\left\|\mQ_k\right\|_{\infty}) \\
        \leq                                   & \frac{1}{1-\gamma} -\alpha \frac{1}{1-\gamma} +\alpha + \alpha \frac{\gamma}{1-\gamma}   \\
        \leq                                   & \frac{1}{1-\gamma}.
    \end{align*}
    The proof is completed by the induction argument. Moreover, since $\mQ^*(s,a)=\E\left[ \sum_{k=1}^{\infty}\gamma^k r_k \mid s,a,\pi^*\right]$, and $|r_k|\leq 1$, we have $|\mQ^*(s,a)| \leq \frac{1}{1-\gamma}$.
\end{proof}

% \begin{lemma}\label{lem:A_bound}
%     For any $\mQ\in\R^{|\gS||\gA|}$, we have \(\left\|\mA_{\mQ}\right\|_{\infty}\leq 1- \alpha(1-\gamma) \).
% \end{lemma}
% \begin{proof}
%     Recalling the definition of $\mA_{\mQ}=\mI+\alpha (\gamma \mP\mPi^{\mQ}-\mI)$, we have
%         \begin{align*}
%         \left\| \mI+\alpha (\gamma \mP\mPi^{\mQ}-\mI) \right\|_{\infty} \leq & 1-\alpha + \alpha \gamma \\
%         =& 1-\alpha (1-\gamma).
%     \end{align*}
%     This completes the proof.
% \end{proof}

\begin{lemma}\label{lem:w_k_naive_bound}
    For $k \in \sN$, we have \(\left\|\vw_k \right\|_{\infty} \leq \frac{2}{1-\gamma}\).
\end{lemma}
\begin{proof}
    Applying triangle inequality to $\vw_k$ in~(\ref{eq:vw_k}) yields
    \begin{align*}
        \left\|\vw_k \right\|_{\infty}\leq & \left\| \hat{\mR}_k-\mR\right\|_{\infty}+\gamma \left\| \mPi^{\mQ^*}\mQ^* \right\|_{\infty}\left\| \hat{\mP}_k-\mP \right\|_{\infty} \\
        \leq                               & \left\|\hat{\mR}_k-\mR \right\|_{\infty}+\frac{\gamma}{1-\gamma} \left\| \hat{\mP}_k-\mP  \right\|_{\infty}                          \\
        \leq                               & 2+ \frac{2\gamma}{1-\gamma}                                                                                                          \\
        =                                  & \frac{2}{1-\gamma},
    \end{align*}
    where the second inequality follows from Lemma~\ref{lem:Q_k_bound} in the~\ref{app:property_Q}.
\end{proof}

\section{Omitted Proofs}\label{app:sec:omitted_proofs}

\subsection{Proof of Lemma~\ref{lem:every_state_action_pair}}\label{app:lem:every_state_action_pair}

\begin{proof}
    Applying the union bound, we have
    \begin{align*}
        \sP\left[ \{ N^{s,a}_m \geq 1 ,\;\forall (s,a)\in\gS\times\gA \}^c \right]  \leq & \sum_{s,a \in\gS\times \gA} \sP\left[ N^{s,a}_m =0\right] \\
        \leq                                                                             & |\gS||\gA| (1-d_{\min})^m                                 \\
        \leq                                                                             & |\gS||\gA| \exp(-md_{\min})                               \\
        \leq                                                                             & \frac{\delta}{2},
    \end{align*}
    where the second inequality follows from the fact that the probability of observing $(s,a)$ is $d(s,a)$ while probability of observing different state-action pair is $1-d(s,a)$, the third inequality holds from the fact that $1-x\leq \exp(-x)$, and the last inequality follows from choice of $m$.
\end{proof}

\subsection{Proof of Lemma~\ref{lem:A^Q_k-bound}}\label{app:lem:A^Q_k-bound}

\begin{proof}
    Note that under event $\gE$ defined in~(\ref{event:every_visit}), we have $\sum_{j=1}^{|\gS|}[\hat{\mP}_k]_{i,j}=1$ for $1 \leq i \leq |\gS||\gA|$. Therefore, we have $\sum^{|\gS||\gA|}_{j=1}[\hat{\mP}\mPi^{\mQ}]_{i,j}=\sum^{|\gS||\gA|}_{j=1} \sum_{h=1}^{|\gS|}[\hat{\mP}_k]_{i,h}[\mPi^{\mQ}]_{h,j}=\sum_{h=1}^{|\gS|} [\hat{\mP}_k]_{i,h} \sum_{j=1}^{|\gS||\gA|}[\mPi^{\mQ}]_{h,j}=1$.

    Hence, we get
    \begin{align*}
             & \left\|(1-\alpha)\mI+\gamma\alpha \hat{\mP}_k\mPi^{\mQ} \right\|_{\infty}                                                \\
        \leq & 1-\alpha + \gamma\alpha\max_{1\leq i \leq |\gS||\gA|}\left(\sum_{j=1}^{|\gS||\gA| } [\hat{\mP}_k\mPi^{\mQ}]_{i,j}\right) \\
        =    & 1-(1-\gamma)\alpha.
    \end{align*}

    This completes the proof.
\end{proof}

\subsection{Proof of Lemma~\ref{lem:inf-norm-bound-P_k-R_k}}\label{app:sec:lem:inf-norm-bound-P_k-R_k}
\begin{proof}
    We will first check the condition in~(\ref{ineq:omega_bound_condition}) to apply Lemma~\ref{lem:omega_bound}. From Lemma~\ref{lem:Q_k_bound}, we have $\left\| \hat{\mP}_k\mPi^{\mQ^*}\mQ^* \right\|_{\infty} \leq \frac{1}{1-\gamma}$. Moreover, noting that $\E\left[ [\hat{\mP}_k \mPi^{\mQ^*}\mQ^*]_{s,a}\right]=[\mP\mPi^{\mQ^*}\mQ^*]_{s,a}$, from Lemma~\ref{lem:binom_tail_bound}, we get, for $1\leq t \leq k$,
    \begin{align*}
             & \sP\left[\left| [\hat{\mP}_k\mPi^{\mQ^*}\mQ^*]_{s,a}  - [\mP\mPi^{\mQ^*}\mQ^*]_{s,a}\right| \geq \eps \middle | N^{s,a}_k = t\right] \\
        \leq & 2\exp(-t\eps^2(1-\gamma)^2/2).
    \end{align*}
    Hence, we can now apply Lemma~\ref{lem:omega_bound}, which yields
    \begin{align}
             & \sP\left[ \left| [\hat{\mP}_k\mPi^{\mQ^*}\mQ^*]_{s,a}-[\mP\mPi^{\mQ^*}\mQ^*]_{s,a}  \right|  \geq \eps  \right]  \nonumber \\
        \leq & 2\exp(-kd(s,a)(1-\gamma)^2\eps^2/4)+\exp(-kd(s,a))\nonumber                                                                \\
        \leq & 3 \exp(-kd(s,a)(1-\gamma)^2\eps^2/4). \label{ineq:hatp-p}
    \end{align}
    The union bound leads to
    \begin{align*}
             & \sP\left[\left\| \hat{\mP}_k\mPi^{\mQ^*}\mQ^*-  \mP\mPi^{\mQ^*}\mQ^*\right\|_{\infty} \geq \eps \right]                          \\
        =    & \sP\left[\max_{s,a\in\gS\times \gA}\left|[\hat{\mP}_k\mPi^{\mQ^*}\mQ^*-  \mP\mPi^{\mQ^*}\mQ^*]_{s,a}\right| \geq \eps \right]    \\
        \leq & \sum_{s,a\in\gS\times\gA} \sP\left[ \left| [\hat{\mP}_k\mPi^{\mQ^*}\mQ^*-  \mP\mPi^{\mQ^*}\mQ^*]_{s,a} \right| \geq \eps \right] \\
        \leq & 3|\gS||\gA| \exp(-kd_{\min}(1-\gamma)^2\eps^2/4),
    \end{align*}
    where the first inequality follows from $  \left\{ \max_{(s,a)\in\gS\times \gA}\left|[\hat{\mP}_k\mPi^{\mQ^*}\mQ^*-  \mP\mPi^{\mQ^*}\mQ^*]_{s,a}\right\| \geq \eps \right\} \subset \cup_{(s,a)\in\gS\times\gA} \left\{\left|[\hat{\mP}_k\mPi^{\mQ^*}\mQ^*-  \mP\mPi^{\mQ^*}\mQ^*]_{s,a}\right| \geq \eps \right\} $, and the last inequality follows from~(\ref{ineq:hatp-p}).
    Furthermore, we can derive the concentration bound for $\hat{\mR}_k$ in the same manner. From Lemma~\ref{lem:binom_tail_bound}, we have
    \begin{align*}
        \sP\left[\left| \hat{r}^{s,a}_k-\mR^{s,a}_k \right|\geq \eps \middle| N^{s,a}_k=t\right] \leq 2 \exp(-t\eps^2/2).
    \end{align*}
    Therefore, from Lemma~\ref{lem:omega_bound}, one gets
    \begin{align*}
        \sP\left[ \left| \hat{r}^{s,a}_k-\mR^{s,a}_k \right|\geq \eps\right] \leq & 2\exp(-kd(s,a)\eps^2/4)  \\
                                                                                  & + \exp(-kd(s,a))         \\
        \leq                                                                      & 3\exp(-kd(s,a)\eps^2/4).
    \end{align*}

    Applying the union bound leads to
    \begin{align*}
        \sP\left[\left\|\hat{\mR}_k-\mR\right\|_{\infty}\geq \eps\right] = & \sP\left[\max_{s,a\in\gS\times \gA} |\hat{r}^{s,a}_k-\mR^{s,a} | \geq \eps  \right] \\
        \leq                                                               & \sum_{s,a\in\gS\times\gA} \sP\left[|\hat{r}^{s,a}_k-\mR^{s,a} | \geq \eps \right]   \\
        \leq                                                               & 3|\gS||\gA| \exp(-kd_{\min}\eps^2/4).
    \end{align*}
    This completes the proof.
\end{proof}

\subsection{Proof of Lemma~\ref{lem:w_concentration_bound}}\label{app:lem:w_concentration_bound}

\begin{proof}
    Applying triangle inequality to the definition of $\vw_k$ in~(\ref{eq:vw_k}) leads to
    \begin{align*}
        \left\|\vw_k \right\|_{\infty}\leq \left\| \hat{\mR}_k-\mR \right\|_{\infty}+\gamma \left\| \hat{\mP}_k\mPi^{\mQ^*}\mQ^*- \mP\mPi^{\mQ^*}\mQ^*\right\|_{\infty}.
    \end{align*}
    Therefore, the following holds:
    \begin{align*}
             & \sP\left[ \left\| \vw_k \right\|_{\infty} < \eps \right]                                                                                                                                                 \\
        \geq & \sP\left[ \left\{  \left\|\hat{\mR}_k-\mR\right\|_{\infty} < \frac{\eps}{2} \right\} \cap\left\{ \left\|  (\hat{\mP}_k- \mP)\mPi^{\mQ^*}\mQ^* \right\|_{\infty} < \frac{\eps}{2\gamma} \right\} \right].
    \end{align*}
    Considering the complement of the event $\{ \left\| \tilde{\vw}_k \right\|_{\infty} < \eps \}^c =\{  \left\| \tilde{\vw}_k \right\|_{\infty} \geq  \eps\} $,
    \begin{align*}
             & \sP\left[  \left\| \vw_k \right\|_{\infty} \geq  \eps \right]                                                                                                                                               \\
        \leq & \sP\left[\left\{  \left\|\hat{\mR}_k-\mR\right\|_{\infty} \geq \frac{\eps}{2} \right\} \cup \left\{ \left\|(\hat{\mP}_k- \mP)\mPi^{\mQ^*}\mQ^* \right\|_{\infty} \geq \frac{\eps}{2\gamma} \right\} \right] \\
        \leq & \sP\left[ \left\|\hat{\mR}_k-\mR\right\|_{\infty} \geq \frac{\eps}{2}\right] + \sP \left[ \left\| (\hat{\mP}_k-\mP)\mPi^{\mQ^*}\mQ^*\right\|_{\infty} \geq \frac{\eps}{2\gamma} \right]                     \\
        \leq & 3|\gS||\gA|\exp(-kd_{\min}\eps^2/16)                                                                                                                                                                        \\
             & + 3|\gS||\gA|\exp(-kd_{\min}(1-\gamma)^2\eps^2/(16\gamma^2))                                                                                                                                                \\
        \leq & 6 |\gS||\gA| \exp(-kd_{\min}(1-\gamma)^2\eps^2/16),
    \end{align*}
    where the second inequality follows from union bound and the third inequality follows from Lemma~\ref{lem:inf-norm-bound-P_k-R_k}.
\end{proof}

\subsection{Proof of Theorem~\ref{thm:sample_complexity_proof}}\label{app:thm:sample_complexity_proof}

\begin{proof}
    Suppose the following event holds, for $k\geq m$:
    \begin{align*}
        W_k := \left\{\left\|\vw_i\right\|_{\infty} \leq \eps^{\prime} ,\; \forall \left\lceil \frac{k-m}{2} \right\rceil +m \leq i \leq k \right\},
    \end{align*}
    The choice of positive constant $\eps^{\prime}$ will be deferred.
    Recursively expanding~(\ref{eq:general_recursion}), we have
    \begin{align*}
        \vx_{k+1} = \prod_{i=m}^k \mA^{\vy_i}_i\vx_m +\alpha \sum_{i=m}^{k-1}\prod_{j=i+1}^{k} \mA^{\vy_j}_j \vw_i +\alpha \vw_k.
    \end{align*}
    Under the event $\gE$ in~(\ref{event:every_visit}), we have $\left\|\mA^{\vy_k}_k\right\|_{\infty} \leq 1-(1-\gamma)\alpha$. Taking infinity norm on the above equation, and applying triangle inequality, we get,
    \begin{align*}
             & \left\|\vx_{k+1} \right\|_{\infty}                                                                                                                 \\
        \leq & \left\|\prod^k_{i=m}\mA^{\vy_i}_i \vx_m \right\|_{\infty}                                                                                          \\
             & +\alpha \sum^{k-1}_{i=m} \prod_{j=i+1}^{k}\left\|\mA^{\vy_j}_j\right\|_{\infty}\left\|\vw_i\right\|_{\infty} +\alpha \left\|\vw_k\right\|_{\infty} \\
        \leq & (1-(1-\gamma)\alpha)^{k-m+1}\left\|\vx_m\right\|_{\infty}                                                                                          \\
             & + \alpha \underbrace{ \sum^{k-1}_{i=m} (1-(1-\gamma)\alpha)^{k-i}\left\|\vw_i\right\|_{\infty}}_{(\star)} + \alpha \left\|\vw_k\right\|_{\infty}   \\
        \leq & (1-(1-\gamma)\alpha)^{k-m+1}\left\|\vx_m\right\|_{\infty}                                                                                          \\
             & + \alpha \sum^{m+\lfloor \frac{k-m}{2} \rfloor}_{i=m} (1-(1-\gamma)\alpha)^{k-i}\left\|\vw_i\right\|_{\infty}                                      \\
             & +\alpha \sum^{k-1}_{i=m+\lceil  \frac{k-m}{2} \rceil} (1-(1-\gamma)\alpha)^{k-i}\left\|\vw_i\right\|_{\infty}+\alpha \left\|\vw_k\right\|_{\infty} \\
        \leq & \underbrace{\frac{2}{1-\gamma }\exp(-(1-\gamma)\alpha(k-m+1))}_{E_1}                                                                               \\
             & + \underbrace{\frac{2}{(1-\gamma)^2} (1-(1-\gamma)\alpha)^{k-m-\lceil \frac{k-m}{2}\rceil }}_{E_2}                                                 \\
             & + \underbrace{\frac{2}{1-\gamma}\eps^{\prime}}_{E_3},
    \end{align*}
    where the third inequality follows from decomposition of $(\star)$ and the last inequality follows from Lemma~\ref{lem:w_k_naive_bound} and the definition of event $W_k$.

    Our aim is to bound the above inequality with $\eps$. One sufficient condition to achieve the bound is to bound each $E_1,E_2$, and $E_3$ with $\eps/3$. First, to bound $E_1$, we need $\exp(-(1-\gamma)\alpha(k-m+1))\leq \frac{\eps(1-\gamma)}{3}$, which is satisfied when
    \begin{align}
        k \geq m + \frac{1}{\alpha(1-\gamma)}\ln\frac{6}{\eps(1-\gamma)}. \label{ineq:E1}
    \end{align}
    Next, to bound $E_2$, we require,
    \begin{align*}
             & \frac{2}{(1-\gamma)^2} (1-(1-\gamma)\alpha)^{k-m-\lfloor \frac{k-m}{2} \rfloor} \\
        \leq & \frac{2}{(1-\gamma)^2} (1-(1-\gamma)\alpha)^{\frac{k-m}{2}}                     \\
        \leq & \frac{2}{(1-\gamma)^2}\exp\left(-\frac{(1-\gamma)\alpha}{2}(k-m)\right)         \\
        \leq & \frac{\eps}{3},
    \end{align*}
    which is satisfied if
    \begin{align}
             & \frac{k-m}{2} \geq \frac{2}{(1-\gamma)\alpha } \ln\left(\frac{6}{\eps(1-\gamma)^2}\right)    \nonumber \\
        \iff & k\geq \frac{4}{(1-\gamma)\alpha} \ln \left(\frac{6}{\eps(1-\gamma)^2}\right)+m. \label{ineq:E2}
    \end{align}
    To bound $E_3$ with $\frac{\eps}{3}$, we require, $\eps^{\prime}\leq \frac{1-\gamma}{6}\eps$.

    Therefore, it is enough to find $k$ such that the event $W_k$ under $\gE$ with $\eps^{\prime}=\frac{1-\gamma}{6}\eps$ to hold with probability at least $1-\frac{\delta}{2}$ from the following relation:
    \begin{align}
        \sP\left[ \left\|\mQ_k-\mQ^* \right\|_{\infty} \leq \eps  \middle | \gE \right] \geq \sP\left[W_k \middle | \gE \right] \geq 1- \frac{\delta}{2}. \label{prob_small_eps}
    \end{align}

    Taking the the complement of the event $W_k$ yields
    \begin{align*}
             & \sP\left[W^c_k \middle | \gE \right]                                                                                                                    \\
        =    & \sP\left[ \cup^k_{i=m+\lceil  \frac{k-m}{2} \rceil} \left\{ \left\| \vw_i \right\|_{\infty} \geq  \frac{1-\gamma}{6}\eps\right\} \middle | \gE \right]  \\= & \frac{1}{\sP[\gE]} \sP\left[ \cup^k_{i=m+\lceil \frac{k-m}{2} \rceil} \left\{ \left\| \vw_i \right\|_{\infty} \geq  \frac{1-\gamma}{6}\eps\right\}  \cap \gE \right] \\
        \leq & \frac{\sP\left[\cup^k_{i=m+\lceil \frac{k-m}{2} \rceil} \left\{ \left\| \vw_i \right\|_{\infty} \geq  \frac{1-\gamma}{6}\eps\right\} \right]}{\sP[\gE]} \\
        \leq & \frac{1}{\sP[\gE]} \sum^{k}_{i=m+\lceil \frac{k-m}{2}\rceil }\sP\left[\left\| \vw_i \right\|_{\infty} \geq  \frac{1-\gamma}{6}\eps\right]               \\
        \leq & \frac{6|\gS||\gA|}{1-\delta/2} \left( k-m+1-\left\lceil \frac{k-m}{2}\right\rceil\right)                                                                \\
             & \times \exp\left(-\left(m+\left\lceil \frac{k-m}{2} \right\rceil\right)d_{\min}(1-\gamma)^4\eps^2/576\right),
    \end{align*}
    where the second equality follows from the law of conditional probability. The last inequality follows from Lemma~\ref{lem:every_state_action_pair} and Lemma~\ref{lem:w_concentration_bound}. For $\sP[W^c_k|\gE]\leq \frac{\delta}{2}$ to hold, the following condition is sufficient:
    \begin{align*}
        k \geq m+2 + \frac{1152}{\eps^2(1-\gamma)^4d_{\min}}   \ln \left( \frac{12k|\gS||\gA|}{\delta(1-\delta/2)} \right).
    \end{align*}
    Noting that $\delta(1-\delta/2)\geq \delta/2$, the following number of samples is sufficient:
    \begin{align}
        k \geq m+2 + \frac{1152}{\eps^2(1-\gamma)^4d_{\min}}\ln \left( \frac{24k|\gS|\gA|}{\delta}\right). \label{ineq:E3}
    \end{align}

    Letting $k$ be the minimum value satisfying the inequalities in~(\ref{ineq:E1}),~(\ref{ineq:E2}) and~(\ref{ineq:E3}), we have $\left\| \mQ_k-\mQ^* \right\|_{\infty}\leq \eps$ with probability at least $1-\frac{\delta}{2}$ under event $\gE$.

    % Letting $S=\{ \left\|\mQ_k-\mQ^*\right\|_{\infty} \leq \eps \}$. Then, we have, 

    % \begin{align*}
    %      & \sP[S |\gE] \geq \sP\left[\cap^k_{i=m+\lceil  \frac{k-m}{2} \rceil} \left\{ \left\| \vw_i \right\|_{\infty} \leq  \frac{1-\gamma}{6}\eps\right\}    \middle | \gE\right] \\
    %     \iff & \sP[S^c\mid \gE] \leq \sP\left[\cup^k_{i=m+\lceil  \frac{k-m}{2} \rceil} \left\{ \left\| \vw_i \right\|_{\infty} \geq  \frac{1-\gamma}{6}\eps\right\}  \middle | \gE\right] \leq \frac{\delta}{2}.
    % \end{align*}

    % Therefore, $\sP[S^c \mid \gE  ] \leq \frac{\delta}{2} $.

    The law of total probability yields
    \begin{align*}
        \sP\left[ \left\|\mQ_k-\mQ^* \right\|_{\infty} \geq \eps \right] = & \sP\left[ \left\{ \left\|\mQ_k-\mQ^* \right\|_{\infty} \geq \eps \right\} \middle| \gE \right] \sP[\gE]      \\
                                                                           & +\sP\left[ \left\{ \left\|\mQ_k-\mQ^* \right\|_{\infty} \geq \eps \right\} \middle| \gE^c \right] \sP[\gE^c] \\
        \leq                                                               & \sP\left[\left\|\mQ_k-\mQ^* \right\|_{\infty} \geq \eps\middle | \gE\right] + \sP[\gE^c]                     \\
        \leq                                                               & \frac{\delta}{2} + \frac{\delta}{2}                                                                          \\
        =                                                                  & \delta.
    \end{align*}
    The second last inequality follows from~(\ref{prob_small_eps}) and Lemma~\ref{lem:every_state_action_pair}.
\end{proof}

\end{document}